\providecommand{\FVQSourceRoot}{}
\ifcsname iflatexml\endcsname\else\expandafter\newif\csname iflatexml\endcsname\fi
\iflatexml
\documentclass{article}
\else
\documentclass{style}

\usepackage[variablett]{lmodern}
\renewcommand{\sfdefault}{optimistic}
\fi

\usepackage{amsmath,amsfonts,bm}

\def\1{\bm{1}}

\DeclareMathAlphabet{\mathsfit}{\encodingdefault}{\sfdefault}{m}{sl}
\SetMathAlphabet{\mathsfit}{bold}{\encodingdefault}{\sfdefault}{bx}{n}

\usepackage{amsmath,amssymb}
\usepackage{booktabs}
\usepackage{multirow}
\usepackage{tabularx}
\usepackage{float}
\usepackage{graphicx}
\usepackage{placeins}
\iflatexml\else\usepackage{fontawesome5}\fi
\usepackage{microtype}
\usepackage{xcolor}
\iflatexml\else
\usepackage{tcolorbox}
\tcbuselibrary{skins}
\fi
\usepackage{hyperref}
\usepackage{url}
\iflatexml
\definecolor{metablue}{HTML}{0064E0}
\usepackage[round,authoryear]{natbib}
\fi


\graphicspath{{fig/}}

\hypersetup{
    colorlinks=true,
    linkcolor=metablue,
    citecolor=metablue,
    urlcolor=metablue,
    pdftitle={Can Discrete Tokens Serve as Semantically Lossless Representations of Visual Signals?},
    pdfauthor={Jialv Zou, Xiaotong Li, Zhengrong Yue, Xu Huang, Yaxin Luo, Wenhao Yan, Haodong Li, Wenyu Liu, Xinggang Wang}
}

\iflatexml
\newenvironment{findingbox}{\begin{quote}}{\end{quote}}
\else
\newtcolorbox{findingbox}{
    colback=black!3,
    colframe=teal!55!black,
    boxrule=0.8pt,
    arc=3pt,
    left=7pt,
    right=7pt,
    top=6pt,
    bottom=6pt,
    before skip=6pt,
    after skip=8pt
}
\fi

\title{Can Discrete Tokens Serve as Semantically Lossless Representations of Visual Signals?}

\iflatexml
\newcommand{\correspondingauthorsymbol}{✉}
\author{Jialv Zou\textsuperscript{1,*,\ensuremath{\diamond}} \quad
Xiaotong Li\textsuperscript{2,*,\ensuremath{\dagger}} \quad
Zhengrong Yue\textsuperscript{3,\ensuremath{\diamond}} \quad
Xu Huang\textsuperscript{4,\ensuremath{\diamond}} \quad
Yaxin Luo\textsuperscript{5,\ensuremath{\diamond}} \quad
Wenhao Yan\textsuperscript{6,\ensuremath{\diamond}} \quad
Haodong Li\textsuperscript{7,\ensuremath{\diamond}} \quad
Wenyu Liu\textsuperscript{1} \quad
Xinggang Wang\textsuperscript{1,\correspondingauthorsymbol}
\\ \textsuperscript{1}Huazhong University of Science and Technology \quad
\textsuperscript{2}Meituan LongCat
\\ \textsuperscript{3}Shanghai Jiao Tong University \quad
\textsuperscript{4}Peking University
\\ \textsuperscript{5}Mohamed bin Zayed University of Artificial Intelligence
\\ \textsuperscript{6}Tsinghua University \quad
\textsuperscript{7}The Chinese University of Hong Kong
\\ {\small\textsuperscript{*}Equal contribution, \textsuperscript{\correspondingauthorsymbol}Corresponding author, \textsuperscript{\ensuremath{\dagger}}Project lead}
\\ {\small\textsuperscript{\ensuremath{\diamond}}Work done during an internship at Meituan LongCat}}
\date{}
\else
\DeclareRobustCommand{\correspondingauthorsymbol}{\faEnvelope[regular]}
\author[1,\,*,\,\ensuremath{\diamond}]{Jialv Zou}
\author[2,\,*,\,\ensuremath{\dagger}]{Xiaotong Li}
\author[3,\,\ensuremath{\diamond}]{Zhengrong Yue}
\author[4,\,\ensuremath{\diamond}]{Xu Huang}
\author[5,\,\ensuremath{\diamond}]{Yaxin Luo}
\author[6,\,\ensuremath{\diamond}]{Wenhao Yan}
\author[7,\,\ensuremath{\diamond}]{Haodong Li}
\author[1]{Wenyu Liu}
\author[1,\,\correspondingauthorsymbol]{Xinggang Wang}

\renewcommand{\affiliationlist}{%
    \affiliationformat[1]{Huazhong University of Science and Technology}\quad
    \affiliationformat[2]{Meituan LongCat}\\
    \affiliationformat[3]{Shanghai Jiao Tong University}\quad
    \affiliationformat[4]{Peking University}\\
    \affiliationformat[5]{Mohamed bin Zayed University of Artificial Intelligence}\\
    \affiliationformat[6]{Tsinghua University}\quad
    \affiliationformat[7]{The Chinese University of Hong Kong}%
}

\contribution[*]{Equal contribution}
\contribution[\correspondingauthorsymbol]{Corresponding author}
\contribution[\ensuremath{\dagger}]{Project lead}
\addtolist[\ensuremath{\diamond}]{Work done during an internship at Meituan LongCat}{\contributionlist}{\contributionformat}{\par}
\fi

\newcommand{\paperabstract}[1]{\gdef\paperabstracttext{#1}}
\paperabstract{
Discrete visual tokens represent vision in a discrete form analogous to language, offering a native interface for multimodal modeling under a shared next-token prediction objective. However, quantization error is widely assumed to cause unavoidable semantic loss, fundamentally limiting the semantic fidelity of discrete visual tokens and positioning continuous visual representations as an intrinsic upper bound. Using multimodal understanding as a proxy for semantic fidelity, we revisit this assumption and find quantization error unreliable, as discrete representations with substantially larger errors can achieve better understanding. Instead, the effective bit budget of discrete indices emerges as a key determinant of downstream performance. However, existing quantizers eventually saturate as additional codebooks contribute progressively less semantic information. As one approach to alleviating this saturation, we introduce Factorized Vector Quantization (FVQ), which quantizes complementary, fixed-dimensional views of the visual feature with symmetric parallel codebooks. It continues to improve at larger bit budgets, ultimately matching continuous visual representations in multimodal understanding. Trained on industrial-scale multimodal data, our model achieves state-of-the-art performance among discrete-representation models, outperforming UniAR and Emu3-Chat by 22.8\% and 59.2\%, respectively, on a nine-metric average. To our knowledge, it is also the first discrete-representation model competitive with strong continuous-representation models such as Qwen3-VL, including on fine-grained OCR tasks. We further identify the key principles and a practical recipe enabling this competitiveness. Together, these results and insights support the potential of discrete visual representations to underpin large-scale native multimodal foundation models.
}

\iflatexml\else\abstract{\paperabstracttext}\fi
\iflatexml\else
\resource{\faGlobe}{Project Website}{https://doctor-james.github.io/FVQ-Page/}
\resource{\faGithub}{Code Repository}{https://github.com/hustvl/FVQ}
\fi

\begin{document}

\maketitle
\iflatexml
\begin{abstract}
\paperabstracttext
\end{abstract}
\begin{description}
\item[Project Website] \url{https://doctor-james.github.io/FVQ-Page/}
\item[Code Repository] \url{https://github.com/hustvl/FVQ}
\end{description}
\fi

\section{Introduction}

Recent multimodal models are moving beyond the conventional pipeline of attaching a standalone vision encoder to a language model through approaches such as early-fusion pretraining~\citep{qwen3.5,team2026kimi} and encoder-free designs~\citep{diao2024unveiling,diao2025evev2,diao2026sensenova}. More ambitiously, recent work pursues the grand unification of multimodal modeling by representing images as discrete tokens analogous to text~\citep{wang2024emu3,geng2025x,team2026longcat}. By bringing vision and language under a shared token interface, this formulation allows multimodal models to inherit the mature architectures, training paradigms, and scaling recipes of LLMs.
Beyond serving as a common modeling interface, discrete visual representations can also function as visual chains of thought~\citep{yang2026context,sun2026vot} or intermediate semantic representations~\citep{wang2026representation}, providing explicit guidance for image generation, spatial understanding, and depth estimation within unified multimodal models.

However, quantizing continuous visual features into a finite codebook inevitably incurs quantization error, which is widely assumed to entail semantic loss. Under this view, discrete representations cannot serve as a semantically faithful proxy for the original visual signal, making continuous visual representations a natural upper bound on downstream performance.

\begin{figure*}[t]
    \centering
    \includegraphics[width=0.98\textwidth]{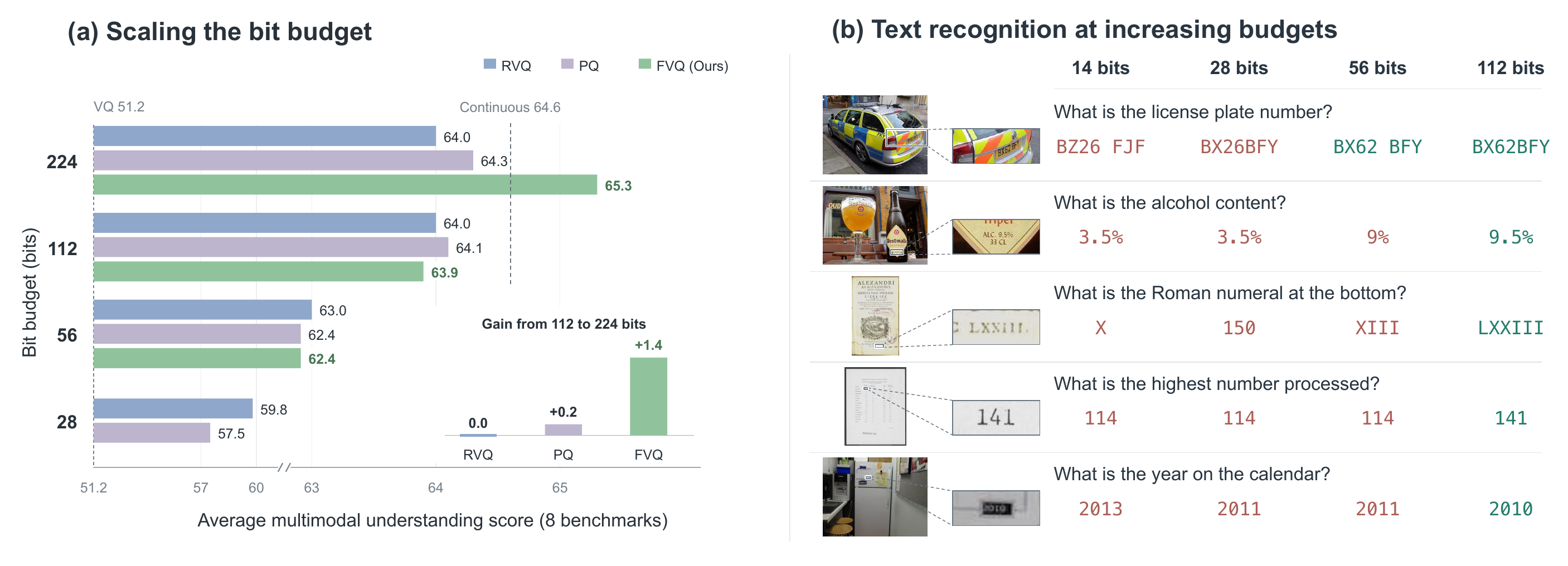}
    \caption{Scaling the bit budget of discrete visual representations. (a) Average multimodal understanding performance improves with the per-patch bit budget. FVQ continues to improve in the high-budget regime and reaches the continuous QwenViT baseline, whereas existing multi-codebook quantizers saturate. (b) Qualitative examples of text recognition at increasing bit budgets. Higher budgets progressively recover the fine-grained textual information required for correct predictions; green and red denote \textcolor[HTML]{237C6A}{correct} and \textcolor[HTML]{AB5D55}{incorrect} answers, respectively.}
    \label{fig:bit_budget_scaling}
\end{figure*}

In this work, we challenge this common assumption. We use multimodal understanding as a proxy for semantic fidelity, since a representation that provides the information needed to answer any image-centric question can be regarded as semantically lossless with respect to the original visual signal~\citep{team2026longcat}. Our experiments reveal that a discrete representation with an order-of-magnitude larger quantization error can nevertheless achieve substantially better understanding performance. This result demonstrates that quantization error is not a reliable measure of semantic fidelity, raising a fundamental question: \emph{What determines the semantic fidelity of discrete visual representations?}

Returning to first principles, raw visual signals, continuous features, and discrete representations are all finite-bit encodings in digital systems, whose nominal capacities can be characterized by their bit budgets, as illustrated in Fig.~\ref{fig:chapter2_overview} (right). The semantic content of raw visual signals can in principle be represented with far fewer bits owing to their high redundancy, suggesting that discrete representations may approach semantic completeness with a sufficiently large bit budget. For discrete representations, the bit budget grows linearly with the number of codebooks $L$ but only logarithmically with the size of each codebook $C$, making multi-codebook quantization a more efficient scaling strategy. To test this hypothesis, we systematically scale the bit budget and observe steady improvements in downstream multimodal understanding, as shown in Fig.~\ref{fig:bit_budget_scaling}(a).

However, existing multi-codebook quantizers face a scaling bottleneck as additional codebooks become less semantically informative. Residual vector quantization (RVQ)~\citep{lee2022autoregressive} assigns later codebooks to progressively smaller residuals, whereas product quantization (PQ)~\citep{jegou2011product,ma2026unitok} partitions the feature into increasingly narrow channel groups. Consequently, the effective semantic capacity fails to keep pace with the nominal bit budget, limiting further scaling.

The observed saturation motivates quantizer designs that continue to provide newly added codebooks with informative views as the bit budget grows. As one approach to alleviating this bottleneck, we introduce Factorized Vector Quantization (FVQ), illustrated in Fig.~\ref{fig:fvq_pipeline}, which constructs complementary, fixed-dimensional views of the same visual feature and quantizes them with symmetric parallel codebooks. Fig.~\ref{fig:bit_budget_scaling}(a) shows that FVQ continues to improve with larger bit budgets after RVQ and PQ saturate, ultimately matching continuous visual representations in multimodal understanding.

Moving beyond small-scale feasibility studies, we train our model on industrial-scale multimodal data. The resulting model achieves competitive performance with flagship multimodal models across a broad suite of benchmarks, demonstrating the data scalability of discrete visual representations and supporting their potential to underpin large-scale native multimodal foundation models.

Our main contributions are summarized as follows.
\begin{itemize}
    \item We show that quantization error does not determine semantic fidelity. Starting from the first-principles observation that digital RGB images are themselves discrete finite-bit signals, we further demonstrate that a sufficiently large effective bit budget enables discrete representations to match continuous ones in downstream understanding.
    \item We introduce Factorized Vector Quantization (FVQ), which provides one approach to alleviating the high-budget saturation observed in existing multi-codebook quantizers. FVQ quantizes complementary, fixed-dimensional views of the same visual feature with symmetric parallel codebooks and sustains gains as the bit budget increases.
    \item We demonstrate the data scalability of our model through industrial-scale multimodal training, achieving state-of-the-art performance among discrete-representation models with relative improvements of 22.8\% over UniAR and 59.2\% over Emu3-Chat on a nine-metric average. To the best of our knowledge, ours is the first discrete-representation model to achieve competitive performance against flagship continuous-representation models such as Qwen3-VL. Beyond these empirical results, we distill the key principles and a practical recipe that make such competitiveness possible, further supporting the potential of discrete visual representations to underpin large-scale native multimodal foundation models.
\end{itemize}

\section{What Determines the Semantics of Discrete Visual Representations?}
\label{sec:semantic_exploration}

\begin{figure}[H]
    \centering
    \includegraphics[width=\textwidth]{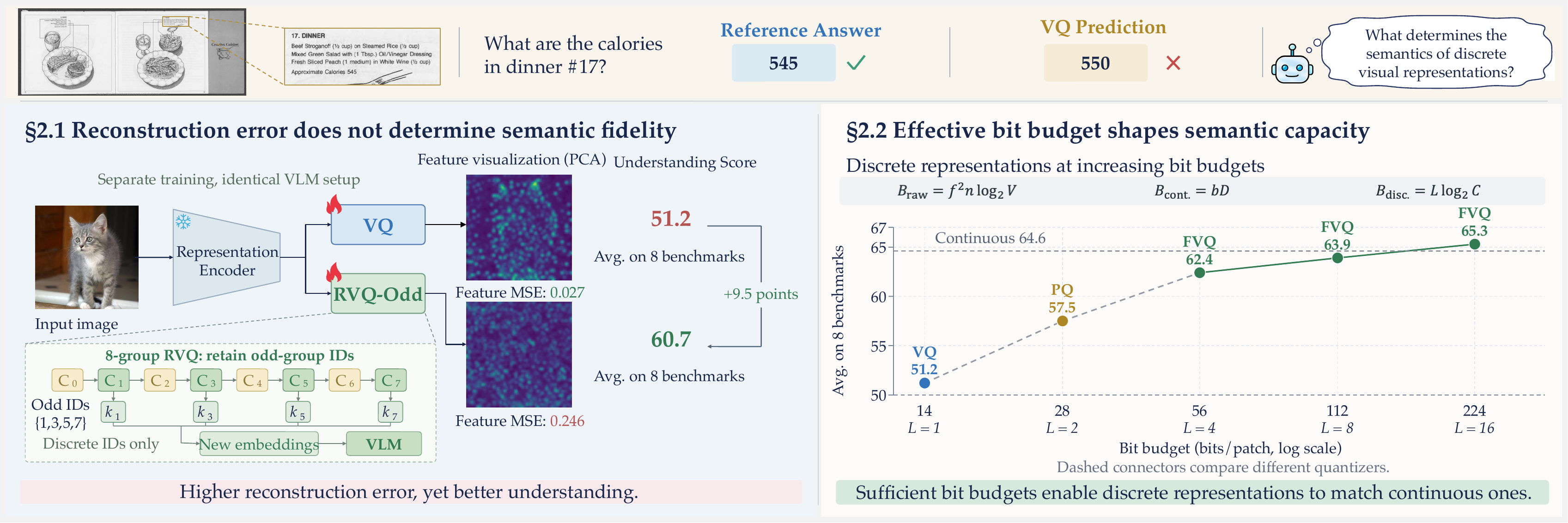}
    \caption{Overview of the analyses in Section~\ref{sec:semantic_exploration}: reconstruction error versus semantic fidelity (left) and bit-budget scaling across representative quantizers (right). $B_{\mathrm{raw}}$, $B_{\mathrm{cont}}$, and $B_{\mathrm{disc}}$ denote per-patch budgets for raw pixels, continuous features, and discrete indices. $f$, $n$, $V$, $b$, $D$, $L$, and $C$ denote patch side length, channels, values per channel, bits per feature dimension, feature dimension, number of codebooks, and codebook size, respectively.}
    \label{fig:chapter2_overview}
\end{figure}

Discrete visual representations provide a shared token interface for vision and language, enabling native multimodal modeling and visual chains of thought while potentially reducing modality conflicts. Their limited semantic fidelity, however, remains a central bottleneck. We therefore investigate what governs the semantic fidelity of discrete visual representations. Section~\ref{sec:quantization_error} tests whether quantization error necessarily implies semantic loss, and Section~\ref{sec:effective_bit_budget} examines effective bit budget and the scaling bottlenecks of existing quantizers. Fig.~\ref{fig:chapter2_overview} summarizes the two analyses.

\subsection{Rethinking Quantization Error and Semantic Fidelity}
\label{sec:quantization_error}

\begin{findingbox}
\textbf{Finding 1: Quantization Error Does Not Determine Semantic Fidelity.} Representations with similar errors can exhibit substantially different understanding performance, whereas those with markedly different errors can retain comparable semantic information.
\end{findingbox}

In this section, we challenge the common assumption that continuous visual representations form an intrinsic upper bound for their discrete counterparts. A prevailing view holds that mapping continuous visual features to a finite codebook inevitably introduces quantization error and that this distortion necessarily entails semantic loss. Under this view, discrete representations are inherently weaker than continuous ones.

However, the VLM operates on discrete indices rather than reconstructed features, so quantization error need not translate directly into reduced semantic fidelity. We compare single-codebook VQ~\citep{oord2017neural} with eight-group RVQ and construct two partial representations by retaining alternating groups. Fig.~\ref{fig:chapter2_overview} (left) highlights the odd-group comparison, while Table~\ref{tab:error_capacity} reports feature-reconstruction MSE on the ImageNet validation set~\citep{deng2009imagenet} alongside downstream multimodal understanding performance for all variants.

\begin{table}[htbp]
    \centering
    \caption{Quantization error and multimodal understanding of discrete QwenViT representations. $C$ and $L$ denote the codebook size and number of selected codebooks, respectively. MSE measures feature-reconstruction error. Avg. is the unweighted mean over the eight benchmarks, with OCRBench divided by 10.}
    \label{tab:error_capacity}
    \setlength{\tabcolsep}{4pt}
    \renewcommand{\arraystretch}{1.12}
    \resizebox{\linewidth}{!}{%
    \begin{tabular}{@{}lccccccccccc@{}}
        \toprule
        \multicolumn{3}{c}{Quantization}
        & \multicolumn{9}{c}{Multimodal Understanding} \\
        \cmidrule(lr){1-3}\cmidrule(lr){4-12}
        Quantizer
        & $C\times L$
        & \shortstack{MSE $\downarrow$}
        & MathVista
        & RealWorldQA
        & MMStar
        & OCRBench
        & TextVQA
        & AI2D
        & ChartQA
        & DocVQA
        & Avg. $\uparrow$ \\
        \midrule
        VQ               & $16{,}384\times1$    & 0.027          & 39.4          & 52.4          & 40.5          & 367          & 51.3          & 67.1          & 54.2          & 68.2          & 51.2 \\
        \midrule
        RVQ (Odd)        & $16{,}384\times4$    & 0.246          & 41.4          & 56.2          & 41.7          & 615          & 66.9          & 75.1          & 63.5          & 79.5          & 60.7 \\
        RVQ (Even)       & $16{,}384\times4$    & 0.028          & 40.3          & 56.2          & 41.7          & 564          & 66.1          & 73.6          & 61.7          & 78.4          & 59.3 \\
        RVQ (All)        & $16{,}384\times8$    & \textbf{0.016} & \textbf{43.7} & \textbf{59.3} & \textbf{43.1} & \textbf{652} & \textbf{71.6} & \textbf{78.8} & \textbf{67.3} & \textbf{82.8} & \textbf{64.0} \\
        \bottomrule
    \end{tabular}
    }
\end{table}

The results reveal a counterintuitive pattern. Although the odd-group variant omits the coarse component captured by group 0 and incurs substantially greater reconstruction error, it markedly outperforms VQ and performs comparably to the even-group variant in understanding performance. Conversely, the even-group variant exhibits reconstruction error comparable to VQ but achieves much stronger understanding. Together, these contrasting cases show that reconstruction error does not determine semantic fidelity, raising a more fundamental question: what governs the semantic capacity of a discrete representation?

\subsection{Bit-Budget Scaling in Discrete Visual Representations}
\label{sec:effective_bit_budget}

\begin{findingbox}
\textbf{Finding 2: Larger Bit Budgets Improve Semantic Capacity When Added Codes Remain Informative.} Increasing the discrete bit budget improves semantic fidelity and downstream multimodal understanding, ultimately enabling discrete representations to match the understanding performance of continuous visual representations.
\end{findingbox}

Returning to first principles, raw visual signals, continuous features, and discrete representations are all finite-bit encodings in digital systems, and their nominal capacities can be characterized by their per-patch bit budgets~\citep{yu2026autoregressive}, as summarized in Fig.~\ref{fig:chapter2_overview} (right). The semantic content of raw visual signals can in principle be represented with far fewer bits owing to their high redundancy. Accordingly, we hypothesize that increasing the discrete bit budget improves semantic fidelity, provided that the additional codes remain semantically informative.

To test this hypothesis, we scale the discrete bit budget and evaluate the resulting representations on downstream multimodal understanding tasks. The discrete bit budget $B_{\mathrm{disc}}=L\log_2 C$ grows linearly with the number of codebooks $L$ but only logarithmically with the codebook size $C$. Larger codebooks are more difficult to optimize and may suffer from poor utilization or codebook collapse. Keeping the codebook size fixed while increasing the number of codebooks therefore provides a more efficient and tractable scaling strategy. We compare RVQ with product quantization (PQ)~\citep{jegou2011product,ma2026unitok}, which partitions each feature into channel groups and quantizes them in parallel. Fig.~\ref{fig:bit_budget_scaling}(a) shows that both methods improve as the bit budget increases and progressively approach the continuous representation. This trend also extends to OCR, where discrete representations often struggle because quantization can discard high-frequency details~\citep{yu2026viq}. Larger budgets progressively recover the fine-grained textual information required for correct predictions, as shown in Fig.~\ref{fig:bit_budget_scaling}(b). Detailed benchmark results are reported in Table~\ref{tab:codebook_scaling}.

Despite this trend, both RVQ and PQ saturate at high bit budgets. Increasing the number of codebooks from 8 to 16 doubles the nominal bit budget but yields only marginal gains in understanding performance. We attribute this saturation to diminishing information per codebook, as later RVQ groups receive weaker residuals and PQ partitions features into increasingly narrow channel groups. The nominal capacity therefore grows without a corresponding increase in effective semantic capacity, showing that additional bits help only when the added codes remain semantically informative. This observation motivates quantizer designs that continue to provide informative representations to newly added codebooks as their number grows. In the next section, we introduce FVQ as one such approach.

\section{From Bit-Budget Scaling to Quantizer Design}
\label{sec:fvq}

\subsection{Overview}

\begin{figure*}[t]
    \centering
    \includegraphics[width=0.95\textwidth]{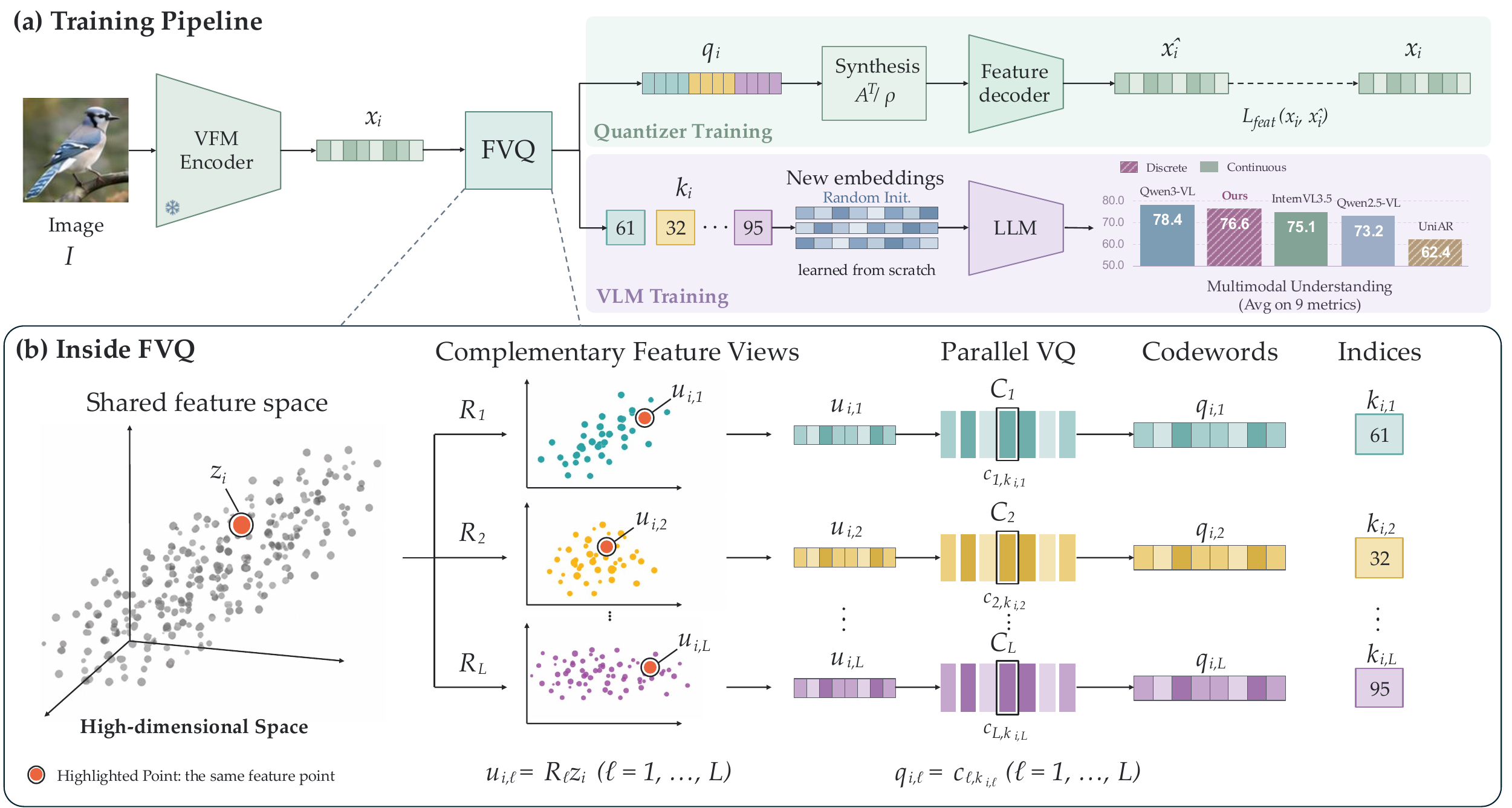}
    \caption{Overview of FVQ. During quantizer training, multiple dense projections produce complementary views of the same visual feature, which are quantized in parallel by independent codebooks. The selected codewords are combined through tight-frame synthesis for feature reconstruction. In the subsequent VLM training stage, the quantizer is frozen, and only the resulting discrete indices are mapped to newly initialized token embeddings for multimodal understanding. The nine-metric average is computed over MMMU, MathVista, MMStar, OCRBench/10, ChartQA, CharXiv-DQ, CharXiv-RQ, DocVQA, and InfoVQA.}
    \label{fig:fvq_pipeline}
\end{figure*}

The preceding analysis shows that a larger nominal bit budget is insufficient when the information available to each codebook diminishes as their number grows. A scalable multi-codebook quantizer should therefore assign symmetric roles to its codebooks, provide each with a fixed-dimensional view derived from the full feature, and collectively preserve the original feature before quantization. Guided by these principles, we introduce \emph{Factorized Vector Quantization} (FVQ) as a concrete realization based on dense tight-frame projections, parallel independent codebooks, and information-preserving synthesis, as illustrated in Fig.~\ref{fig:fvq_pipeline}.

\subsection{Factorized Vector Quantization}

\paragraph{Complementary Tight-Frame Projections.}

Let $\mathbf{z}_i\in\mathbb{R}^{D}$ denote the projected VFM feature of the $i$-th visual patch. FVQ maps $\mathbf{z}_i$ into $L$ fixed-dimensional dense views. Each projection $\mathbf{R}_\ell\in\mathbb{R}^{d\times D}$ is dense over all $D$ input channels, allowing every codebook to receive information from the full feature while keeping its input dimension $d$ unchanged as $L$ grows. Unlike PQ, increasing $L$ adds another view rather than dividing the feature into progressively narrower channel groups. We stack these projections into an analysis operator $\mathbf{A}$ and constrain them as follows:
\begin{equation}
    \begin{aligned}
        \mathbf{u}_{i,\ell}
        &=\mathbf{R}_\ell\mathbf{z}_i,
        \quad \mathbf{R}_\ell\in\mathbb{R}^{d\times D},\\
        \mathbf{A}
        &=\begin{bmatrix}
            \mathbf{R}_1^{\top} & \cdots & \mathbf{R}_L^{\top}
        \end{bmatrix}^{\top}\in\mathbb{R}^{Ld\times D},\\
        \mathbf{A}^{\top}\mathbf{A}
        &=\rho\mathbf{I}_{D},
        \quad \rho=\frac{Ld}{D}.
    \end{aligned}
    \label{eq:fvq_factorization}
\end{equation}
Dense projections provide global views of the same feature, but independently sampled projections do not guarantee that these views preserve the feature collectively and may weight different directions unevenly. We therefore impose the tight-frame condition $\mathbf{A}^{\top}\mathbf{A}=\rho\mathbf{I}_D$. This makes the aggregate transform isotropic, such that $\|\mathbf{A}\mathbf{z}_i\|_2^2=\rho\|\mathbf{z}_i\|_2^2$. Every feature direction is scaled equally, and no nonzero direction is discarded by all views simultaneously. The projected views therefore preserve the feature geometry collectively before quantization.

For $Ld\geq D$, we construct $\mathbf{A}$ by applying reduced QR decomposition to a fixed Gaussian matrix and scaling the resulting column-orthonormal matrix by $\sqrt{\rho}$. The projections remain fixed throughout training, providing stable input spaces for the corresponding codebooks.

\paragraph{Symmetric Parallel Quantization.}

Each projected view $\mathbf{u}_{i,\ell}$ is quantized independently using a codebook $\mathcal{C}_\ell=\{\mathbf{c}_{\ell,j}\}_{j=1}^{C}$, where $\mathbf{c}_{\ell,j}\in\mathbb{R}^{d}$ and all codebooks contain $C$ entries. FVQ performs nearest-neighbor lookup in parallel:
\begin{equation}
    k_{i,\ell}
    =\arg\min_{j\in\{1,\ldots,C\}}
    \left\|\mathbf{u}_{i,\ell}-\mathbf{c}_{\ell,j}\right\|_2^2.
    \label{eq:fvq_quantization}
\end{equation}
Here, $k_{i,\ell}$ is the selected discrete ID and $\mathbf{c}_{\ell,k_{i,\ell}}$ is the corresponding codeword. Each visual patch is represented by the tuple $(k_{i,1},\ldots,k_{i,L})$, yielding the nominal bit budget $B=L\log_2 C$ illustrated in Fig.~\ref{fig:chapter2_overview} (right). Independent parallel quantization places all codebooks in symmetric roles, allowing this capacity to grow with $L$ without introducing the residual hierarchy of RVQ.

\paragraph{Tight-Frame Synthesis.}

To reconstruct the projected feature $\mathbf{z}_i$ during tokenizer training, we combine the selected codewords using the tight-frame synthesis operator $\rho^{-1}\mathbf{A}^{\top}$. Let $\mathbf{u}_i=[\mathbf{u}_{i,1}^{\top},\ldots,\mathbf{u}_{i,L}^{\top}]^{\top}=\mathbf{A}\mathbf{z}_i$ denote the stacked views before quantization. Then
\begin{equation}
    \begin{aligned}
        \hat{\mathbf{z}}_i
        &=\frac{1}{\rho}\sum_{\ell=1}^{L}
        \mathbf{R}_\ell^{\top}\mathbf{c}_{\ell,k_{i,\ell}},\\
        \frac{1}{\rho}\mathbf{A}^{\top}\mathbf{u}_i
        &=\frac{1}{\rho}\mathbf{A}^{\top}\mathbf{A}\mathbf{z}_i
        =\mathbf{z}_i.
    \end{aligned}
    \label{eq:fvq_synthesis}
\end{equation}
The second line follows directly from $\mathbf{A}^{\top}\mathbf{A}=\rho\mathbf{I}_D$ and shows that the analysis--synthesis transform exactly recovers $\mathbf{z}_i$ before quantization. The factorization itself therefore introduces no reconstruction error. All distortion arises from replacing the projected views with their nearest codewords.

\paragraph{Properties of FVQ.}

FVQ implements the three design choices described above. First, parallel independent quantization assigns symmetric roles to all codebooks, avoiding the progressively weaker inputs induced by a residual hierarchy. Second, dense projections provide each codebook with a fixed-dimensional view derived from the full feature, so the information available to an individual codebook does not structurally diminish as $L$ increases. Third, the tight-frame analysis--synthesis pair makes these views collectively information-preserving before quantization. Together, these choices enable the nominal bit budget to scale with $L$ without structurally reducing the information available to newly added codebooks.

\section{Experiments}
\label{sec:experiments}

\subsection{Implementation Details}
\label{sec:implementation_details}

We report two training configurations: an industrial-scale configuration for the main experiments and a shared configuration based primarily on the LLaVA-OneVision data~\citep{li2024llava} for the exploratory experiments in Section~\ref{sec:semantic_exploration} and the ablation studies.

\paragraph{Model Architecture.}
Our main model pairs Qwen3-8B~\citep{yang2025qwen3} with an any-resolution visual tokenizer constructed by applying 16-codebook FVQ to Qwen3 ViT~\citep{bai2025qwen3vl}. To preserve a native next-token prediction setup with minimal modality-specific modifications to the LLM, we use standard 1D RoPE~\citep{su2024roformer} in the LLM and omit the DeepStack connections from Qwen3 ViT. Unless otherwise specified, the exploratory and ablation experiments use Qwen3-1.7B and vary the quantization configuration according to the experiment.

\paragraph{Training Schedule.}
Both configurations follow four stages. Quantizer training learns the visual tokenizer by reconstructing frozen visual features. Alignment stage freezes both the tokenizer and the LLM and optimizes the newly initialized visual token embeddings and merger to align the discrete IDs with the language embedding space. Pre-training and supervised fine-tuning both keep the tokenizer frozen and jointly optimize the LLM, visual token embeddings, and merger, using context lengths of 8K and 16K and maximum image resolutions of 1024px and 2048px, respectively.

\paragraph{Training Data.}
For the main experiments, quantizer training, alignment, pre-training, and supervised fine-tuning use 100B, 10B, 60B, and 30B tokens, respectively, drawn from mixtures covering STEM, OCR, interleaved image-text sequences, image captioning and text-only data. For the exploratory and ablation experiments, we train the quantizer on 50B tokens, use approximately 4B tokens for alignment, and adopt LLaVA-OneVision~\citep{li2024llava} for pre-training and supervised fine-tuning.

\paragraph{Evaluation.}
We evaluate general perception and reasoning on MMMU~\citep{yue2024mmmu}, its more challenging and robust extension MMMU-Pro~\citep{yue2025mmmu}, MMStar~\citep{chen2024mmstar}, and RealWorldQA~\citep{xai2024realworldqa}. We assess mathematical reasoning on MathVista~\citep{lu2024mathvista}. For text recognition and text-centric understanding, we use OCRBench~\citep{liu2024ocrbench}, OCRBench V2~\citep{fu2026ocrbench}, and TextVQA~\citep{singh2019textvqa}; for document and infographic understanding, we use DocVQA~\citep{mathew2021docvqa}, InfoVQA~\citep{mathew2022infographicvqa}, and OmniDocBench~\citep{ouyang2025omnidocbench}, which evaluates document parsing across diverse PDF document types. We assess diagram and chart understanding on AI2D~\citep{kembhavi2016ai2d} with mask, ChartQA~\citep{masry2022chartqa}, and CharXiv~\citep{wang2024charxiv}.

\subsection{Main Results}
\label{sec:main_results}

\paragraph{Bit-Budget Scaling.}
We compare discrete quantizers with the original continuous Qwen3 ViT representation under identical VLM training and evaluation settings. Table~\ref{tab:codebook_scaling} reports the complete benchmark results underlying the scaling trend in Fig.~\ref{fig:bit_budget_scaling}(a). Increasing the number of codebooks substantially improves RVQ and PQ over single-codebook methods, but their gains diminish at high budgets. FVQ, by contrast, continues to improve with larger budgets and ultimately matches the average understanding performance of continuous QwenViT.
\begin{table}[htbp]
    \centering
    \caption{Bit-budget scaling of discrete QwenViT representations under the shared experimental setting. All discrete variants quantize the same features with codebook size $C=16{,}384$. $L$ is the number of codebooks and $B=L\log_2 C$ is the nominal per-patch bit budget. Bold indicates the best discrete result. Continuous denotes the original QwenViT representation.}
    \label{tab:codebook_scaling}
    \setlength{\tabcolsep}{4pt}
    \renewcommand{\arraystretch}{1.12}
    \resizebox{\linewidth}{!}{%
    \begin{tabular}{@{}lccccccccccc@{}}
        \toprule
        \multicolumn{3}{c}{Quantization}
        & \multicolumn{9}{c}{Multimodal Understanding} \\
        \cmidrule(lr){1-3}\cmidrule(lr){4-12}
        Quantizer
        & $L$
        & $B$
        & MathVista
        & RealWorldQA
        & MMStar
        & OCRBench
        & TextVQA
        & AI2D
        & ChartQA
        & DocVQA
        & Avg. $\uparrow$ \\
        \midrule
         VQ~\citep{oord2017neural} & 1  & 14  & 39.4          & 52.4          & 40.5          & 367          & 51.3          & 67.1          & 54.2          & 68.2          & 51.2 \\
         SimVQ~\citep{zhu2025addressing} & 1  & 14  & 37.9          & 49.7          & 39.1          & 204          & 32.2          & 63.9          & 41.9          & 49.9          & 41.9 \\
        \midrule
         \multirow{4}{*}{RVQ~\citep{lee2022autoregressive}}
                 & 2  & 28  & 41.4          & 54.8          & 42.7          & 563          & 65.8          & 74.3          & 64.4          & 79.0          & 59.8 \\
                 & 4  & 56  & 43.2          & 58.7          & 43.1          & 625          & 69.7          & 78.2          & 66.8          & 81.7          & 63.0 \\
                 & 8  & 112 & 43.7          & 59.3          & 43.1          & 652          & 71.6          & 78.8          & 67.3          & 82.8          & 64.0 \\
                 & 16 & 224 & 44.4          & 56.2          & 43.2          & 649          & \textbf{72.3} & 80.2          & 68.6          & 82.5          & 64.0 \\
        \midrule
         \multirow{4}{*}{PQ~\citep{ma2026unitok}}
                 & 2  & 28  & 41.1          & 55.9          & 41.4          & 489          & 60.8          & 73.6          & 61.6          & 76.8          & 57.5 \\
                 & 4  & 56  & 44.6          & 58.0          & 42.7          & 603          & 68.0          & 77.8          & 66.6          & 81.2          & 62.4 \\
                 & 8  & 112 & \textbf{46.0} & 57.3          & \textbf{45.3} & 623          & 70.1          & \textbf{80.3} & 68.7          & 82.4          & 64.1 \\
                 & 16 & 224 & 45.1          & 57.6          & 43.9          & 642          & 71.6          & 79.4          & 68.8          & 83.6          & 64.3 \\
        \midrule
         \multirow{3}{*}{\textbf{FVQ (Ours)}}
                 & 4  & 56  & 45.8          & 57.1          & 42.7          & 600          & 68.3          & 78.1          & 66.3          & 81.1          & 62.4 \\
                 & 8  & 112 & 44.5          & 57.6          & 44.2          & 658          & 69.7          & 79.3          & 68.2          & 81.7          & 63.9 \\
                 & 16 & 224 & 45.8          & \textbf{61.7} & 44.3          & \textbf{663} & 71.9          & 79.0          & \textbf{69.2} & \textbf{84.0} & \textbf{65.3} \\
        \midrule
        Continuous & -- & -- & 45.7          & 60.1          & 45.8          & 676          & 71.3          & 76.7          & 67.4          & 81.9          & 64.6 \\
        \bottomrule
    \end{tabular}
    }
\end{table}

\paragraph{STEM \& General.}
Table~\ref{tab:main_results} shows that our model achieves state-of-the-art performance among discrete-representation models while remaining competitive with strong continuous-representation models. It matches Qwen3-VL-8B on MathVista and achieves the best RealWorldQA result, demonstrating that discrete visual representations can support strong reasoning and general understanding under industrial-scale training.
\begin{table}[htbp]
    \centering
    \caption{Comparison on STEM and general multimodal understanding benchmarks. For a fair comparison, results marked with \textsuperscript{*} are obtained by us under the same evaluation setting as our model.}
    \label{tab:main_results}
    \setlength{\tabcolsep}{3pt}
    \renewcommand{\arraystretch}{1.12}
    \small
    \begin{tabular*}{\linewidth}{@{\extracolsep{\fill}}lccccc@{}}
        \toprule
        Method & MMMU & MMMU-Pro & MathVista & RealWorldQA & MMStar \\
        \midrule
        \multicolumn{6}{l}{\textcolor{gray}{\textit{Continuous}}} \\
        \addlinespace[2pt]
        LLaVA-OV-1.5-8B~\citep{an2025llava} & 55.4 & 31.3 & 69.6 & 68.1 & 67.7 \\
        LLaVA-OV-2-8B~\citep{an2026llava} & -- & -- & -- & 69.7 & 64.8 \\
        InternVL3-8B~\citep{zhu2025internvl3} & 62.7 & -- & 71.6 & 70.8 & 68.2 \\
        InternVL3.5-8B~\citep{wang2025internvl3} & 73.4 & -- & 78.4 & 67.5 & 69.3 \\
        Qwen2.5-VL-7B~\citep{Qwen2.5-VL} & 58.6 & 38.3 & 68.2 & -- & 63.9 \\
        Qwen3-VL-8B~\citep{bai2025qwen3vl} & 69.6 & 55.9 & 77.2 & 71.5 & 70.9 \\
        \midrule
        \multicolumn{6}{l}{\textcolor{gray}{\textit{Discrete}}} \\
        \addlinespace[2pt]
        Emu3-Chat-8B~\citep{wang2024emu3} & 31.6 & -- & 47.6\textsuperscript{*} & 57.4 & 46.5\textsuperscript{*} \\
        Tar-7B~\citep{han2026vision} & 39.0 & -- & -- & -- & -- \\
        ViQ-7B~\citep{yu2026viq} & 49.1 & -- & -- & -- & 54.2 \\
        UniAR-8B~\citep{peng2026unified} & 44.3 & 22.7\textsuperscript{*} & 50.8\textsuperscript{*} & 68.5 & 56.7\textsuperscript{*} \\
        \textbf{Ours-8B} & 64.3 & 40.8 & 77.4 & 72.0 & 68.5 \\
        \bottomrule
    \end{tabular*}
\end{table}

\paragraph{OCR \& Doc \& Chart.}
Quantization is commonly thought to discard the high-frequency details required for OCR, document, and chart understanding, causing discrete-representation models to lag substantially behind continuous-representation models~\citep{yu2026viq}. Table~\ref{tab:main_results_ocr} shows that our model substantially narrows this gap, achieving state-of-the-art performance among discrete-representation models, performing comparably to Qwen3-VL-8B on OmniDocBench and DocVQA, and surpassing the continuous-representation models on both CharXiv tasks. These results suggest that weak fine-grained visual understanding is not intrinsic to discretization but instead reflects the information bottleneck imposed by an insufficient effective bit budget. By expanding this budget, FVQ alleviates the bottleneck and substantially improves fine-grained visual understanding, challenging the view that discrete visual representations are inherently ill-suited to OCR.
\begin{table}[htbp]
    \centering
    \caption{Comparison of OCR, chart, and document understanding performance. For a fair comparison, results marked with \textsuperscript{*} are obtained by us under the same evaluation setting as our model.}
    \label{tab:main_results_ocr}
    \setlength{\tabcolsep}{2pt}
    \renewcommand{\arraystretch}{1.12}
    \footnotesize
    \resizebox{\linewidth}{!}{%
    \begin{tabular}{@{}lccccccc@{}}
        \toprule
        Method & OCRBench & OCRBenchV2 (en/zh) & ChartQA & CharXiv (DQ/RQ) & DocVQA & InfoVQA & OmniDocBench (en/zh)$\downarrow$ \\
        \midrule
        \multicolumn{8}{l}{\textcolor{gray}{\textit{Continuous}}} \\
        \addlinespace[2pt]
        LLaVA-OV-1.5-8B~\citep{an2025llava} & 829 & -- & 86.5 & 70.9/-- & 95.0 & 78.4 & -- \\
        LLaVA-OV-2-8B~\citep{an2026llava} & 782 & -- & 85.9 & -- & 95.2 & 74.4 & -- \\
        InternVL3-8B~\citep{zhu2025internvl3} & 880 & -- & 86.6 & 73.6/37.6 & 92.7 & 76.8 & -- \\
        InternVL3.5-8B~\citep{wang2025internvl3} & 832 & -- & 86.7 & 72.2/44.4 & 92.3 & 76.2 & -- \\
        Qwen2.5-VL-7B~\citep{Qwen2.5-VL} & 864 & 56.3/57.2 & 87.3 & 73.9/42.5 & 95.7 & 82.6 & 0.308/0.398 \\
        Qwen3-VL-8B~\citep{bai2025qwen3vl} & 896 & 65.4/61.2 & 89.6 & 83.0/46.4 & 96.1 & 83.1 & 0.170/0.264 \\
        \midrule
        \multicolumn{8}{l}{\textcolor{gray}{\textit{Discrete}}} \\
        \addlinespace[2pt]
        Emu3-Chat-8B~\citep{wang2024emu3} & 687 & -- & 68.6 & 36.8\textsuperscript{*}/14.1\textsuperscript{*} & 76.3 & 42.7\textsuperscript{*} & -- \\
        X-Omni-7B~\citep{geng2025x} & 704 & -- & -- & -- & 88.6 & -- & -- \\
        ViQ-7B~\citep{yu2026viq} & 711 & -- & 72.8 & -- & 88.9 & 55.3 & -- \\
        UniAR-8B~\citep{peng2026unified} & 833 & 43.0\textsuperscript{*}/48.6\textsuperscript{*} & 75.9 & 63.3\textsuperscript{*}/26.8\textsuperscript{*} & 91.4 & 68.9\textsuperscript{*} & 0.482\textsuperscript{*}/0.434\textsuperscript{*} \\
        \textbf{Ours-8B} & 867 & 58.7/59.5 & 83.9 & 86.1/49.3 & 95.4 & 77.7 & 0.179/0.249 \\
        \bottomrule
    \end{tabular}
    }
\end{table}

\subsection{Visualization of Fine-Grained Visual Understanding}
\label{sec:ocr_hard_cases}

\begin{figure}[htbp]
    \centering
    \includegraphics[width=0.98\textwidth]{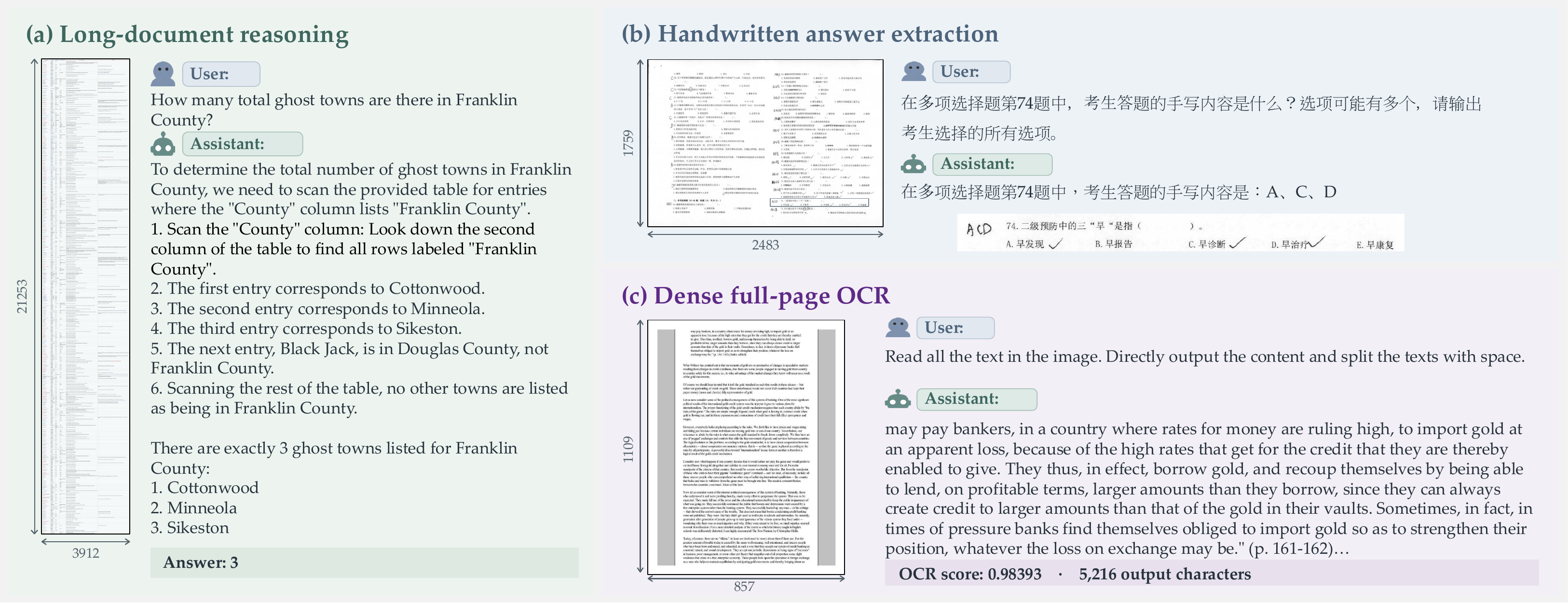}
    \caption{Challenging cases for fine-grained visual understanding: (a) reasoning over an extremely high-resolution long document, (b) extracting irregular handwritten answers, and (c) long-form transcription of a dense full page.}
    \label{fig:ocr_hard_cases}
\end{figure}

Fig.~\ref{fig:ocr_hard_cases} presents three cases that probe fine-grained visual understanding beyond standard OCR. They require locating and aggregating sparse entries in a long table at an extreme resolution of $21{,}253\times3{,}912$ pixels, extracting small and irregular handwritten selections from a dense scanned examination sheet, and sustaining accurate transcription across a densely typeset page with more than 5,000 output characters. Our model succeeds in all three settings, challenging the view that discrete visual representations are intrinsically incapable of fine-grained visual understanding.

\FloatBarrier
\subsection{Ablation Studies}
\label{sec:ablation_studies}

Unless otherwise specified, all ablation studies are conducted using FVQ with eight codebooks under the shared setting described in Section~\ref{sec:implementation_details}.

\paragraph{Generalization across Vision Encoders.}
We examine whether the bit-budget finding generalizes beyond QwenViT by comparing single-codebook VQ with eight-codebook FVQ on features from MoonViT-v2~\citep{team2026kimi} and SigLIP2~\citep{tschannen2025siglip} under the same setting. Table~\ref{tab:ablation_moonvit_quantization} shows substantial average gains for both encoders, supporting our view that effective bit budget is a general determinant of semantic capacity across visual representations.

\begin{table}[htbp]
    \centering
    \caption{Comparison of VQ~\citep{oord2017neural} and FVQ across vision encoders. $B$ denotes the per-patch bit budget; $B=14$ and $B=112$ correspond to single-codebook VQ and eight-codebook FVQ, respectively.}
    \label{tab:ablation_moonvit_quantization}
    \setlength{\tabcolsep}{4pt}
    \renewcommand{\arraystretch}{0.90}
    \small
    \resizebox{\linewidth}{!}{%
    \begin{tabular}{@{}llccccccccc@{}}
        \toprule
        Encoder & $B$ & MathVista & RealWorldQA & MMStar & OCRBench & TextVQA & AI2D & ChartQA & DocVQA & Avg. $\uparrow$ \\
        \midrule
        \multirow{2}{*}{MoonViT-v2} & 14  & 34.9 & 47.5 & 36.5 & 198 & 36.7 & 63.3 & 27.9 & 43.0 & 38.7 \\
                                     & 112 & 44.6 & 54.1 & 39.9 & 608 & 62.1 & 78.4 & 69.5 & 79.7 & 61.1 \\
        \midrule
        \multirow{2}{*}{SigLIP2} & 14  & 38.6 & 52.7 & 43.5 & 287 & 47.4 & 68.3 & 44.6 & 33.7 & 44.7 \\
                                  & 112 & 40.0 & 56.3 & 43.9 & 499 & 64.7 & 75.6 & 57.1 & 43.3 & 53.9 \\
        \bottomrule
    \end{tabular}}
\end{table}

\paragraph{Comparison under LLaVA-OneVision SFT.}
To isolate the effect of FVQ from industrial-scale training, we retain only the alignment stage and then fine-tune the model on LLaVA-OneVision SFT data, omitting the industrial-scale pre-training and SFT data used for our main model. Table~\ref{tab:ablation_llava_ov_comparison} shows that FVQ achieves the strongest overall performance among the compared methods.

\begin{table}[htbp]
    \centering
    \caption{Comparison under LLaVA-OneVision SFT. LLaVA-OneVision uses Qwen2-7B~\citep{li2024llava}; all other models use Qwen2.5-7B. LLaVA-OneVision results are from its paper, and the other baselines are reported by ViQ~\citep{yu2026viq}. Bold indicates the best discrete result.}
    \label{tab:ablation_llava_ov_comparison}
    \setlength{\tabcolsep}{3pt}
    \renewcommand{\arraystretch}{0.90}
    \small
    \resizebox{\linewidth}{!}{%
    \begin{tabular}{@{}lccccccc@{}}
        \toprule
        Method & MMMU & MathVista & RealWorldQA & MMStar & OCRBench & ChartQA & DocVQA \\
        \midrule
        \multicolumn{8}{l}{\textcolor{gray}{\textit{Continuous}}} \\
        \addlinespace[2pt]
        LLaVA-OneVision~\citep{li2024llava} & 48.8 & 63.2 & 66.3 & 61.7 & -- & 80.0 & 87.5 \\
        SigLIP2-g~\citep{tschannen2025siglip} & 48.3 & -- & -- & 57.2 & 671 & 72.8 & 75.0 \\
        AIMv2-H~\citep{fini2025multimodal} & 48.2 & -- & -- & 55.2 & 687 & 72.5 & 80.1 \\
        InternViT-2.5-6B~\citep{chen2024expanding} & 48.1 & -- & -- & 55.3 & 757 & 77.4 & 85.7 \\
        \midrule
        \multicolumn{8}{l}{\textcolor{gray}{\textit{Discrete}}} \\
        \addlinespace[2pt]
        ViQ~\citep{yu2026viq} & 49.1 & -- & -- & 54.2 & 711 & 72.8 & 88.9 \\
        \textbf{Ours} & \textbf{53.1} & 60.2 & 65.2 & \textbf{58.5} & \textbf{791} & \textbf{81.6} & \textbf{90.3} \\
        \bottomrule
    \end{tabular}}
\end{table}

\section{Conclusion}

We revisit the semantic fidelity of discrete visual tokens and show that feature reconstruction error does not reliably predict downstream understanding. Instead, effective bit budget emerges as a key determinant of semantic capacity. However, existing multi-codebook quantizers eventually saturate as additional codebooks contribute diminishing semantic information. As one approach to alleviating this bottleneck, FVQ combines complementary projections with symmetric parallel codebooks and sustains gains at larger bit budgets. Moving beyond small-scale feasibility studies, our industrial-scale model achieves state-of-the-art performance among discrete-representation models and remains competitive with strong continuous-representation models, including on fine-grained OCR and document understanding. These results support discrete tokens as a scalable interface for large-scale native multimodal foundation models.

\FloatBarrier
\bibliography{\FVQSourceRoot references}

\begin{thebibliography}{68}
\providecommand{\natexlab}[1]{#1}
\providecommand{\url}[1]{\texttt{#1}}
\expandafter\ifx\csname urlstyle\endcsname\relax
  \providecommand{\doi}[1]{doi: #1}\else
  \providecommand{\doi}{doi: \begingroup \urlstyle{rm}\Url}\fi

\bibitem[An et~al.(2025)An, Xie, Yang, Zhang, Zhao, Cheng, Wang, Xu, Chen, Zhu,
  et~al.]{an2025llava}
Xiang An, Yin Xie, Kaicheng Yang, Wenkang Zhang, Xiuwei Zhao, Zheng Cheng,
  Yirui Wang, Songcen Xu, Changrui Chen, Didi Zhu, et~al.
\newblock {LLaVA-OneVision-1.5}: Fully open framework for democratized
  multimodal training.
\newblock \emph{arXiv preprint arXiv:2509.23661}, 2025.

\bibitem[An et~al.(2026)An, Xie, Tang, Yan, Tan, Zhu, Chen, Zhao, Qin, Yang,
  et~al.]{an2026llava}
Xiang An, Yin Xie, Feilong Tang, Yunyao Yan, Huajie Tan, Didi Zhu, Changrui
  Chen, Xiuwei Zhao, Bin Qin, Kaicheng Yang, et~al.
\newblock {LLaVA-OneVision-2}: Towards next-generation perceptual intelligence.
\newblock \emph{arXiv preprint arXiv:2605.25979}, 2026.

\bibitem[Bai et~al.(2025{\natexlab{a}})Bai, Cai, Chen, Chen, Chen, Cheng, Deng,
  Ding, Gao, Ge, et~al.]{bai2025qwen3vl}
Shuai Bai, Yuxuan Cai, Ruizhe Chen, Keqin Chen, Xionghui Chen, Zesen Cheng,
  Lianghao Deng, Wei Ding, Chang Gao, Chunjiang Ge, et~al.
\newblock {Qwen3-VL} technical report.
\newblock \emph{arXiv preprint arXiv:2511.21631}, 2025{\natexlab{a}}.

\bibitem[Bai et~al.(2025{\natexlab{b}})Bai, Chen, Liu, Wang, Ge, Song, Dang,
  Wang, Wang, Tang, Zhong, Zhu, Yang, Li, Wan, Wang, Ding, Fu, Xu, Ye, Zhang,
  Xie, Cheng, Zhang, Yang, Xu, and Lin]{Qwen2.5-VL}
Shuai Bai, Keqin Chen, Xuejing Liu, Jialin Wang, Wenbin Ge, Sibo Song, Kai
  Dang, Peng Wang, Shijie Wang, Jun Tang, Humen Zhong, Yuanzhi Zhu, Mingkun
  Yang, Zhaohai Li, Jianqiang Wan, Pengfei Wang, Wei Ding, Zheren Fu, Yiheng
  Xu, Jiabo Ye, Xi~Zhang, Tianbao Xie, Zesen Cheng, Hang Zhang, Zhibo Yang,
  Haiyang Xu, and Junyang Lin.
\newblock {Qwen2.5-VL} technical report.
\newblock \emph{arXiv preprint arXiv:2502.13923}, 2025{\natexlab{b}}.

\bibitem[{Chameleon Team}(2024)]{team2024chameleon}
{Chameleon Team}.
\newblock Chameleon: Mixed-modal early-fusion foundation models.
\newblock \emph{arXiv preprint arXiv:2405.09818}, 2024.

\bibitem[Chen et~al.(2024{\natexlab{a}})Chen, Li, Dong, Zhang, Zang, Chen,
  Duan, Wang, Qiao, Lin, et~al.]{chen2024mmstar}
Lin Chen, Jinsong Li, Xiaoyi Dong, Pan Zhang, Yuhang Zang, Zehui Chen, Haodong
  Duan, Jiaqi Wang, Yu~Qiao, Dahua Lin, et~al.
\newblock Are we on the right way for evaluating large vision-language models?
\newblock \emph{Advances in Neural Information Processing Systems},
  37:\penalty0 27056--27087, 2024{\natexlab{a}}.

\bibitem[Chen et~al.(2025)Chen, Wu, Liu, Pan, Liu, Xie, Yu, and
  Ruan]{chen2025janus}
Xiaokang Chen, Zhiyu Wu, Xingchao Liu, Zizheng Pan, Wen Liu, Zhenda Xie,
  Xingkai Yu, and Chong Ruan.
\newblock {Janus-Pro}: Unified multimodal understanding and generation with
  data and model scaling.
\newblock \emph{arXiv preprint arXiv:2501.17811}, 2025.

\bibitem[Chen et~al.(2024{\natexlab{b}})Chen, Wang, Cao, Liu, Gao, Cui, Zhu,
  Ye, Tian, Liu, et~al.]{chen2024expanding}
Zhe Chen, Weiyun Wang, Yue Cao, Yangzhou Liu, Zhangwei Gao, Erfei Cui, Jinguo
  Zhu, Shenglong Ye, Hao Tian, Zhaoyang Liu, et~al.
\newblock Expanding performance boundaries of open-source multimodal models
  with model, data, and test-time scaling.
\newblock \emph{arXiv preprint arXiv:2412.05271}, 2024{\natexlab{b}}.

\bibitem[Deng et~al.(2025)Deng, Zhu, Li, Gou, Li, Wang, Zhong, Yu, Nie, Song,
  et~al.]{deng2025emerging}
Chaorui Deng, Deyao Zhu, Kunchang Li, Chenhui Gou, Feng Li, Zeyu Wang, Shu
  Zhong, Weihao Yu, Xiaonan Nie, Ziang Song, et~al.
\newblock Emerging properties in unified multimodal pretraining.
\newblock \emph{arXiv preprint arXiv:2505.14683}, 2025.

\bibitem[Deng et~al.(2009)Deng, Dong, Socher, Li, Li, and
  Fei-Fei]{deng2009imagenet}
Jia Deng, Wei Dong, Richard Socher, Li-Jia Li, Kai Li, and Li~Fei-Fei.
\newblock {ImageNet}: A large-scale hierarchical image database.
\newblock In \emph{2009 IEEE conference on computer vision and pattern
  recognition}, pages 248--255. IEEE, 2009.

\bibitem[Diao et~al.(2024)Diao, Cui, Li, Wang, Lu, and Wang]{diao2024unveiling}
Haiwen Diao, Yufeng Cui, Xiaotong Li, Yueze Wang, Huchuan Lu, and Xinlong Wang.
\newblock Unveiling encoder-free vision-language models.
\newblock \emph{Advances in Neural Information Processing Systems},
  37:\penalty0 52545--52567, 2024.

\bibitem[Diao et~al.(2025)Diao, Li, Cui, Wang, Deng, Pan, Wang, Lu, and
  Wang]{diao2025evev2}
Haiwen Diao, Xiaotong Li, Yufeng Cui, Yueze Wang, Haoge Deng, Ting Pan, Wenxuan
  Wang, Huchuan Lu, and Xinlong Wang.
\newblock {EVEv2}: Improved baselines for encoder-free vision-language models.
\newblock In \emph{2025 IEEE/CVF International Conference on Computer Vision
  (ICCV)}, pages 21014--21025. IEEE, 2025.

\bibitem[Diao et~al.(2026)Diao, Wu, Deng, Wang, Bai, Wu, Fan, Ye, Tong, Fan,
  et~al.]{diao2026sensenova}
Haiwen Diao, Penghao Wu, Hanming Deng, Jiahao Wang, Shihao Bai, Silei Wu,
  Weichen Fan, Wenjie Ye, Wenwen Tong, Xiangyu Fan, et~al.
\newblock {SenseNova-U1}: Unifying multimodal understanding and generation with
  neo-unify architecture.
\newblock \emph{arXiv preprint arXiv:2605.12500}, 2026.

\bibitem[Fini et~al.(2025)Fini, Shukor, Li, Dufter, Klein, Haldimann,
  Aitharaju, da~Costa, B{\'e}thune, Gan, et~al.]{fini2025multimodal}
Enrico Fini, Mustafa Shukor, Xiujun Li, Philipp Dufter, Michal Klein, David
  Haldimann, Sai Aitharaju, Victor G~Turrisi da~Costa, Louis B{\'e}thune, Zhe
  Gan, et~al.
\newblock Multimodal autoregressive pre-training of large vision encoders.
\newblock In \emph{2025 IEEE/CVF Conference on Computer Vision and Pattern
  Recognition (CVPR)}, pages 9641--9654. IEEE, 2025.

\bibitem[Fu et~al.(2026)Fu, Kuang, Song, Huang, Yang, Li, Zhu, Luo, Wang, Lu,
  et~al.]{fu2026ocrbench}
Ling Fu, Zhebin Kuang, Jiajun Song, Mingxin Huang, Biao Yang, Yuzhe Li, Linghao
  Zhu, Qidi Luo, Xinyu Wang, Hao Lu, et~al.
\newblock {OCRBench V2}: An improved benchmark for evaluating large multimodal
  models on visual text localization and reasoning.
\newblock \emph{Advances in Neural Information Processing Systems}, 38, 2026.

\bibitem[Geng et~al.(2025)Geng, Wang, Ma, Li, Rao, Gu, Zhong, Lu, Hu, Zhang,
  et~al.]{geng2025x}
Zigang Geng, Yibing Wang, Yeyao Ma, Chen Li, Yongming Rao, Shuyang Gu, Zhao
  Zhong, Qinglin Lu, Han Hu, Xiaosong Zhang, et~al.
\newblock {X-Omni}: Reinforcement learning makes discrete autoregressive image
  generative models great again.
\newblock \emph{arXiv preprint arXiv:2507.22058}, 2025.

\bibitem[Han et~al.(2026)Han, Chen, Zhao, Wang, Zhao, Yang, He, Yue, and
  Jiang]{han2026vision}
Jiaming Han, Hao Chen, Yang Zhao, Hanyu Wang, Qi~Zhao, Ziyan Yang, Hao He,
  Xiangyu Yue, and Lu~Jiang.
\newblock Vision as a dialect: Unifying visual understanding and generation via
  text-aligned representations.
\newblock \emph{Advances in Neural Information Processing Systems},
  38:\penalty0 158430--158459, 2026.

\bibitem[J{\'e}gou et~al.(2011)J{\'e}gou, Douze, and Schmid]{jegou2011product}
Herv{\'e} J{\'e}gou, Matthijs Douze, and Cordelia Schmid.
\newblock Product quantization for nearest neighbor search.
\newblock \emph{IEEE Transactions on Pattern Analysis and Machine
  Intelligence}, 33\penalty0 (1):\penalty0 117--128, 2011.
\newblock \doi{10.1109/TPAMI.2010.57}.

\bibitem[Kembhavi et~al.(2016)Kembhavi, Salvato, Kolve, Seo, Hajishirzi, and
  Farhadi]{kembhavi2016ai2d}
Aniruddha Kembhavi, Mike Salvato, Eric Kolve, Minjoon Seo, Hannaneh Hajishirzi,
  and Ali Farhadi.
\newblock A diagram is worth a dozen images.
\newblock In \emph{European conference on computer vision}, pages 235--251.
  Springer, 2016.

\bibitem[{Kimi Team} et~al.(2026){Kimi Team}, Bai, Bai, Bao, Cai, Cai, Cao,
  Cao, Chai, Charles, et~al.]{team2026kimi}
{Kimi Team}, Tongtong Bai, Yifan Bai, Yiping Bao, Jianfeng Cai, Xinyuan Cai,
  Peizhou Cao, Yuxuan Cao, Ziwei Chai, Y~Charles, et~al.
\newblock {Kimi K3}: Open frontier intelligence.
\newblock \emph{arXiv preprint arXiv:2607.24653}, 2026.

\bibitem[Lee et~al.(2022)Lee, Kim, Kim, Cho, and Han]{lee2022autoregressive}
Doyup Lee, Chiheon Kim, Saehoon Kim, Minsu Cho, and Wook-Shin Han.
\newblock Autoregressive image generation using residual quantization.
\newblock In \emph{Proceedings of the IEEE/CVF Conference on Computer Vision
  and Pattern Recognition}, pages 11523--11532, 2022.

\bibitem[Li et~al.(2024)Li, Zhang, Guo, Zhang, Li, Zhang, Zhang, Zhang, Li,
  Liu, et~al.]{li2024llava}
Bo~Li, Yuanhan Zhang, Dong Guo, Renrui Zhang, Feng Li, Hao Zhang, Kaichen
  Zhang, Peiyuan Zhang, Yanwei Li, Ziwei Liu, et~al.
\newblock {LLaVA-OneVision}: Easy visual task transfer.
\newblock \emph{arXiv preprint arXiv:2408.03326}, 2024.

\bibitem[Liu et~al.(2024)Liu, Li, Huang, Yang, Yu, Li, Yin, Liu, Jin, and
  Bai]{liu2024ocrbench}
Yuliang Liu, Zhang Li, Mingxin Huang, Biao Yang, Wenwen Yu, Chunyuan Li,
  Xu-Cheng Yin, Cheng-Lin Liu, Lianwen Jin, and Xiang Bai.
\newblock {OCRBench}: On the hidden mystery of {OCR} in large multimodal
  models.
\newblock \emph{Science China Information Sciences}, 67\penalty0 (12):\penalty0
  220102, 2024.

\bibitem[Loshchilov and Hutter(2017)]{loshchilov2017decoupled}
Ilya Loshchilov and Frank Hutter.
\newblock Decoupled weight decay regularization.
\newblock \emph{arXiv preprint arXiv:1711.05101}, 2017.

\bibitem[Lu et~al.(2024)Lu, Bansal, Xia, Liu, Li, Hajishirzi, Cheng, Chang,
  Galley, and Gao]{lu2024mathvista}
Pan Lu, Hritik Bansal, Tony Xia, Jiacheng Liu, Chunyuan Li, Hannaneh
  Hajishirzi, Hao Cheng, Kai-Wei Chang, Michel Galley, and Jianfeng Gao.
\newblock {MathVista}: Evaluating mathematical reasoning of foundation models
  in visual contexts.
\newblock In \emph{International Conference on Learning Representations},
  volume 2024, pages 23439--23554, 2024.

\bibitem[Luo et~al.(2025)Luo, Yang, Dou, Wang, Liu, Dai, Qiao, and
  Zhu]{luo2025monointernvl}
Gen Luo, Xue Yang, Wenhan Dou, Zhaokai Wang, Jiawen Liu, Jifeng Dai, Yu~Qiao,
  and Xizhou Zhu.
\newblock Mono-{InternVL}: Pushing the boundaries of monolithic multimodal
  large language models with endogenous visual pre-training.
\newblock In \emph{Proceedings of the IEEE/CVF Conference on Computer Vision
  and Pattern Recognition}, pages 24960--24971, 2025.

\bibitem[Ma et~al.(2026{\natexlab{a}})Ma, Jiang, Wu, Yang, Yu, Yuan, Peng, and
  Qi]{ma2026unitok}
Chuofan Ma, Yi~Jiang, Junfeng Wu, Jihan Yang, Xin Yu, Zehuan Yuan, Bingyue
  Peng, and Xiaojuan Qi.
\newblock {UniTok}: A unified tokenizer for visual generation and
  understanding.
\newblock \emph{Advances in Neural Information Processing Systems},
  38:\penalty0 129274--129297, 2026{\natexlab{a}}.

\bibitem[Ma et~al.(2026{\natexlab{b}})Ma, Long, Chen, Zhang, Zhao, Zhang, and
  Ye]{ma2026drae}
Tianren Ma, Lin Long, Chuyan Chen, Mu~Zhang, Junbo Zhao, Tong Zhang, and
  Qixiang Ye.
\newblock {dRAE}: Representation autoencoder with hyper-spherical codes.
\newblock \emph{arXiv preprint arXiv:2607.22148}, 2026{\natexlab{b}}.

\bibitem[Ma et~al.(2025)Ma, Liu, Chen, Liu, Wu, Wu, Pan, Xie, Zhang, Yu,
  et~al.]{ma2025janusflow}
Yiyang Ma, Xingchao Liu, Xiaokang Chen, Wen Liu, Chengyue Wu, Zhiyu Wu, Zizheng
  Pan, Zhenda Xie, Haowei Zhang, Xingkai Yu, et~al.
\newblock {JanusFlow}: Harmonizing autoregression and rectified flow for
  unified multimodal understanding and generation.
\newblock In \emph{2025 IEEE/CVF Conference on Computer Vision and Pattern
  Recognition (CVPR)}, pages 7739--7751. IEEE, 2025.

\bibitem[Masry et~al.(2022)Masry, Tan, Joty, Hoque, et~al.]{masry2022chartqa}
Ahmed Masry, Jia~Qing Tan, Shafiq Joty, Enamul Hoque, et~al.
\newblock {ChartQA}: A benchmark for question answering about charts with
  visual and logical reasoning.
\newblock In \emph{Findings of the association for computational linguistics:
  ACL 2022}, pages 2263--2279, 2022.

\bibitem[Mathew et~al.(2021)Mathew, Karatzas, and Jawahar]{mathew2021docvqa}
Minesh Mathew, Dimosthenis Karatzas, and CV~Jawahar.
\newblock {DocVQA}: A dataset for {VQA} on document images.
\newblock In \emph{2021 IEEE Winter Conference on Applications of Computer
  Vision (WACV)}, pages 2199--2208. IEEE, 2021.

\bibitem[Mathew et~al.(2022)Mathew, Bagal, Tito, Karatzas, Valveny, and
  Jawahar]{mathew2022infographicvqa}
Minesh Mathew, Viraj Bagal, Rub{\`e}n Tito, Dimosthenis Karatzas, Ernest
  Valveny, and CV~Jawahar.
\newblock {InfographicVQA}.
\newblock In \emph{2022 IEEE/CVF Winter Conference on Applications of Computer
  Vision (WACV)}, pages 2582--2591. IEEE, 2022.

\bibitem[{Meituan LongCat Team} et~al.(2026){Meituan LongCat Team}, Xiao, Wang,
  Li, Zhang, Peng, Yu, Yang, Yan, Sun, et~al.]{team2026longcat}
{Meituan LongCat Team}, Bin Xiao, Chao Wang, Chengjiang Li, Chi Zhang, Chong
  Peng, Hang Yu, Hao Yang, Haonan Yan, Haoze Sun, et~al.
\newblock {LongCat-Next}: Lexicalizing modalities as discrete tokens.
\newblock \emph{arXiv preprint arXiv:2603.27538}, 2026.

\bibitem[Mentzer et~al.(2024)Mentzer, Minnen, Agustsson, and
  Tschannen]{mentzer2024finite}
Fabian Mentzer, David Minnen, Eirikur Agustsson, and Michael Tschannen.
\newblock Finite scalar quantization: {VQ-VAE} made simple.
\newblock In \emph{International Conference on Learning Representations}, pages
  51772--51783, 2024.

\bibitem[Ouyang et~al.(2025)Ouyang, Qu, Zhou, Zhu, Zhang, Lin, Wang, Zhao,
  Jiang, Zhao, et~al.]{ouyang2025omnidocbench}
Linke Ouyang, Yuan Qu, Hongbin Zhou, Jiawei Zhu, Rui Zhang, Qunshu Lin, Bin
  Wang, Zhiyuan Zhao, Man Jiang, Xiaomeng Zhao, et~al.
\newblock {OmniDocBench}: Benchmarking diverse {PDF} document parsing with
  comprehensive annotations.
\newblock In \emph{2025 IEEE/CVF Conference on Computer Vision and Pattern
  Recognition (CVPR)}, pages 24838--24848. IEEE, 2025.

\bibitem[Peng et~al.(2026)Peng, Meng, Cai, Zhuang, Yang, Fang, Wu, Lin, Wu, and
  Bai]{peng2026unified}
Wujian Peng, Lingchen Meng, Yuxuan Cai, Xianwei Zhuang, Yuhuan Yang, Rongyao
  Fang, Chenfei Wu, Junyang Lin, Zuxuan Wu, and Shuai Bai.
\newblock Unified multimodal autoregressive modeling with shared
  context---visual tokenizer is key to unification.
\newblock \emph{arXiv preprint arXiv:2606.18249}, 2026.

\bibitem[{Qwen Team}(2026)]{qwen3.5}
{Qwen Team}.
\newblock {Qwen3.5}: Towards native multimodal agents, February 2026.
\newblock \url{https://qwen.ai/blog?id=qwen3.5}.

\bibitem[Radford et~al.(2021)Radford, Kim, Hallacy, Ramesh, Goh, Agarwal,
  Sastry, Askell, Mishkin, Clark, et~al.]{radford2021learning}
Alec Radford, Jong~Wook Kim, Chris Hallacy, Aditya Ramesh, Gabriel Goh,
  Sandhini Agarwal, Girish Sastry, Amanda Askell, Pamela Mishkin, Jack Clark,
  et~al.
\newblock Learning transferable visual models from natural language
  supervision.
\newblock In \emph{International conference on machine learning}, pages
  8748--8763. PMLR, 2021.

\bibitem[Shi et~al.(2025)Shi, Luo, Ge, Yang, Shan, and Wang]{shi2025scalable}
Fengyuan Shi, Zhuoyan Luo, Yixiao Ge, Yujiu Yang, Ying Shan, and Limin Wang.
\newblock Scalable image tokenization with index backpropagation quantization.
\newblock In \emph{2025 IEEE/CVF International Conference on Computer Vision
  (ICCV)}, pages 16037--16046. IEEE, 2025.

\bibitem[Singh et~al.(2019)Singh, Natarajan, Shah, Jiang, Chen, Batra, Parikh,
  and Rohrbach]{singh2019textvqa}
Amanpreet Singh, Vivek Natarajan, Meet Shah, Yu~Jiang, Xinlei Chen, Dhruv
  Batra, Devi Parikh, and Marcus Rohrbach.
\newblock Towards {VQA} models that can read.
\newblock In \emph{2019 IEEE/CVF Conference on Computer Vision and Pattern
  Recognition (CVPR)}, pages 8309--8318. IEEE, 2019.

\bibitem[Su et~al.(2024)Su, Ahmed, Lu, Pan, Bo, and Liu]{su2024roformer}
Jianlin Su, Murtadha Ahmed, Yu~Lu, Shengfeng Pan, Wen Bo, and Yunfeng Liu.
\newblock Roformer: Enhanced transformer with rotary position embedding.
\newblock \emph{Neurocomputing}, 568:\penalty0 127063, 2024.

\bibitem[Sun et~al.(2026)Sun, Liao, Luo, Deng, Zhang, Wang, Yang, Fan, and
  Huang]{sun2026vot}
Jingxiang Sun, Chao Liao, Zhengxiong Luo, Chaorui Deng, Chen-lin Zhang, Junke
  Wang, Ceyuan Yang, Haoqi Fan, and Weilin Huang.
\newblock {VoT}: Vision-of-thought for unified multimodal representation
  alignment.
\newblock \emph{arXiv preprint arXiv:2609.07815}, 2026.

\bibitem[{Thinking Machines Lab}(2026)]{thinkingmachines2026interactionmodels}
{Thinking Machines Lab}.
\newblock Interaction models: A scalable approach to human--{AI} collaboration.
\newblock \emph{Thinking Machines Lab: Connectionism}, May 2026.
\newblock \doi{10.64434/tml.20260511}.
\newblock \url{https://thinkingmachines.ai/blog/interaction-models/}.

\bibitem[Tschannen et~al.(2025)Tschannen, Gritsenko, Wang, Naeem,
  Alabdulmohsin, Parthasarathy, Evans, Beyer, Xia, Mustafa,
  et~al.]{tschannen2025siglip}
Michael Tschannen, Alexey Gritsenko, Xiao Wang, Muhammad~Ferjad Naeem, Ibrahim
  Alabdulmohsin, Nikhil Parthasarathy, Talfan Evans, Lucas Beyer, Ye~Xia, Basil
  Mustafa, et~al.
\newblock {SigLIP 2}: Multilingual vision-language encoders with improved
  semantic understanding, localization, and dense features.
\newblock \emph{arXiv preprint arXiv:2502.14786}, 2025.

\bibitem[van~den Oord et~al.(2017)van~den Oord, Vinyals, and
  Kavukcuoglu]{oord2017neural}
Aaron van~den Oord, Oriol Vinyals, and Koray Kavukcuoglu.
\newblock Neural discrete representation learning.
\newblock In \emph{Advances in Neural Information Processing Systems},
  volume~30, 2017.

\bibitem[Wang et~al.(2025)Wang, Gao, Gu, Pu, Cui, Wei, Liu, Jing, Ye, Shao,
  et~al.]{wang2025internvl3}
Weiyun Wang, Zhangwei Gao, Lixin Gu, Hengjun Pu, Long Cui, Xingguang Wei,
  Zhaoyang Liu, Linglin Jing, Shenglong Ye, Jie Shao, et~al.
\newblock {InternVL3.5}: Advancing open-source multimodal models in
  versatility, reasoning, and efficiency.
\newblock \emph{arXiv preprint arXiv:2508.18265}, 2025.

\bibitem[Wang et~al.(2024{\natexlab{a}})Wang, Zhang, Luo, Sun, Cui, Wang,
  Zhang, Wang, Li, Yu, et~al.]{wang2024emu3}
Xinlong Wang, Xiaosong Zhang, Zhengxiong Luo, Quan Sun, Yufeng Cui, Jinsheng
  Wang, Fan Zhang, Yueze Wang, Zhen Li, Qiying Yu, et~al.
\newblock {Emu3}: Next-token prediction is all you need.
\newblock \emph{arXiv preprint arXiv:2409.18869}, 2024{\natexlab{a}}.

\bibitem[Wang et~al.(2026)Wang, Lin, Yang, Zhao, Xiao, He, Zhao, Ding, Wang,
  Wang, et~al.]{wang2026representation}
Yuqing Wang, Zhijie Lin, Ceyuan Yang, Yang Zhao, Fei Xiao, Hao He, Qi~Zhao,
  Zihan Ding, Fuyun Wang, Shuai Wang, et~al.
\newblock Representation forcing for bottleneck-free unified multimodal models.
\newblock \emph{arXiv preprint arXiv:2605.31604}, 2026.

\bibitem[Wang et~al.(2024{\natexlab{b}})Wang, Xia, He, Chen, Liu, Zhu, Liang,
  Wu, Liu, Malladi, et~al.]{wang2024charxiv}
Zirui Wang, Mengzhou Xia, Luxi He, Howard Chen, Yitao Liu, Richard Zhu, Kaiqu
  Liang, Xindi Wu, Haotian Liu, Sadhika Malladi, et~al.
\newblock {CharXiv}: Charting gaps in realistic chart understanding in
  multimodal {LLMs}.
\newblock \emph{Advances in Neural Information Processing Systems},
  37:\penalty0 113569--113697, 2024{\natexlab{b}}.

\bibitem[Wu et~al.(2025)Wu, Zhang, Chen, Tang, Li, Fang, Zhu, Xie, Yin, Yi,
  et~al.]{wu2025vila}
Yecheng Wu, Zhuoyang Zhang, Junyu Chen, Haotian Tang, Dacheng Li, Yunhao Fang,
  Ligeng Zhu, Enze Xie, Hongxu Yin, Li~Yi, et~al.
\newblock {VILA-U}: A unified foundation model integrating visual understanding
  and generation.
\newblock In \emph{International Conference on Learning Representations},
  volume 2025, pages 93620--93638, 2025.

\bibitem[{xAI}(2024)]{xai2024realworldqa}
{xAI}.
\newblock {RealWorldQA}: A benchmark for real-world spatial understanding.
\newblock \url{https://huggingface.co/datasets/xai-org/RealworldQA}, 2024.
\newblock Accessed: 2025-04-26.

\bibitem[Xie et~al.(2025)Xie, Mao, Bai, Zhang, Wang, Lin, Gu, Chen, Yang, and
  Shou]{xie2025show}
Jinheng Xie, Weijia Mao, Zechen Bai, David~Junhao Zhang, Weihao Wang,
  Kevin~Qinghong Lin, Yuchao Gu, Zhijie Chen, Zhenheng Yang, and Mike~Zheng
  Shou.
\newblock {Show-o}: One single transformer to unify multimodal understanding
  and generation.
\newblock In \emph{International Conference on Learning Representations},
  volume 2025, pages 28240--28264, 2025.

\bibitem[Yang et~al.(2025)Yang, Li, Yang, Zhang, Hui, Zheng, Yu, Gao, Huang,
  Lv, et~al.]{yang2025qwen3}
An~Yang, Anfeng Li, Baosong Yang, Beichen Zhang, Binyuan Hui, Bo~Zheng, Bowen
  Yu, Chang Gao, Chengen Huang, Chenxu Lv, et~al.
\newblock {Qwen3} technical report.
\newblock \emph{arXiv preprint arXiv:2505.09388}, 2025.

\bibitem[Yang et~al.(2026)Yang, Lin, Zhao, Xiao, He, Zhao, Deng, Li, Ding, Guo,
  et~al.]{yang2026context}
Ceyuan Yang, Zhijie Lin, Yang Zhao, Fei Xiao, Hao He, Qi~Zhao, Chaorui Deng,
  Kunchang Li, Zihan Ding, Yuwei Guo, et~al.
\newblock Context unrolling in omni models.
\newblock \emph{arXiv preprint arXiv:2604.21921}, 2026.

\bibitem[Yu et~al.(2024)Yu, Lezama, Gundavarapu, Versari, Sohn, Minnen, Cheng,
  Birodkar, Gupta, Gu, Hauptmann, Gong, Yang, Essa, Ross, and
  Jiang]{yu2024language}
Lijun Yu, Jos\'e Lezama, Nitesh~B. Gundavarapu, Luca Versari, Kihyuk Sohn,
  David Minnen, Yong Cheng, Vighnesh Birodkar, Agrim Gupta, Xiuye Gu,
  Alexander~G. Hauptmann, Boqing Gong, Ming-Hsuan Yang, Irfan Essa, David~A.
  Ross, and Lu~Jiang.
\newblock Language model beats diffusion---tokenizer is key to visual
  generation.
\newblock In \emph{International Conference on Learning Representations}, 2024.

\bibitem[Yu et~al.(2026{\natexlab{a}})Yu, Liu, He, Zhang, Liu, Chen, and
  Chen]{yu2026autoregressive}
Qihang Yu, Qihao Liu, Ju~He, Xinyang Zhang, Yang Liu, Liang-Chieh Chen, and
  Xi~Chen.
\newblock Autoregressive image generation with masked bit modeling.
\newblock \emph{arXiv preprint arXiv:2602.09024}, 2026{\natexlab{a}}.

\bibitem[Yu et~al.(2026{\natexlab{b}})Yu, Liu, Yang, Dong, Qian, Lu, Hu, and
  Rao]{yu2026viq}
Xumin Yu, Zuyan Liu, Zhenyu Yang, Yuhao Dong, Shengsheng Qian, Jiwen Lu, Han
  Hu, and Yongming Rao.
\newblock {ViQ}: Text-aligned visual quantized representations at any
  resolution.
\newblock \emph{arXiv preprint arXiv:2606.27313}, 2026{\natexlab{b}}.

\bibitem[Yuan et~al.(2026{\natexlab{a}})Yuan, Qu, Qian, Chen, Tang, Sun, Chu,
  Zhang, Wang, Cai, and Li]{yuan2025video}
Zhenlong Yuan, Xiangyan Qu, Chengxuan Qian, Rui Chen, Jing Tang, Lei Sun,
  Xiangxiang Chu, Dapeng Zhang, Yiwei Wang, Yujun Cai, and Shuo Li.
\newblock {Video-STAR}: Reinforcing open-vocabulary action recognition with
  tools.
\newblock In \emph{The Fourteenth International Conference on Learning
  Representations}, 2026{\natexlab{a}}.

\bibitem[Yuan et~al.(2026{\natexlab{b}})Yuan, Wang, Zhang, Cui, Chen, Tang,
  Sun, Yu, Qian, Chu, Li, and Zhou]{yuan2026if}
Zhenlong Yuan, Yue Wang, Dapeng Zhang, Kejin Cui, Rui Chen, Jing Tang, Lei Sun,
  Hongwei Yu, Chengxuan Qian, Xiangxiang Chu, Shuo Li, and Yuyin Zhou.
\newblock What if agents could imagine? reinforcing open-vocabulary {HOI}
  comprehension through generation.
\newblock \emph{Advances in Neural Information Processing Systems},
  2026{\natexlab{b}}.

\bibitem[Yue et~al.(2024)Yue, Ni, Zhang, Zheng, Liu, Zhang, Stevens, Jiang,
  Ren, Sun, et~al.]{yue2024mmmu}
Xiang Yue, Yuansheng Ni, Kai Zhang, Tianyu Zheng, Ruoqi Liu, Ge~Zhang, Samuel
  Stevens, Dongfu Jiang, Weiming Ren, Yuxuan Sun, et~al.
\newblock {MMMU}: A massive multi-discipline multimodal understanding and
  reasoning benchmark for expert {AGI}.
\newblock In \emph{Proceedings of the IEEE/CVF conference on computer vision
  and pattern recognition}, pages 9556--9567, 2024.

\bibitem[Yue et~al.(2025)Yue, Zheng, Ni, Wang, Zhang, Tong, Sun, Yu, Zhang,
  Sun, et~al.]{yue2025mmmu}
Xiang Yue, Tianyu Zheng, Yuansheng Ni, Yubo Wang, Kai Zhang, Shengbang Tong,
  Yuxuan Sun, Botao Yu, Ge~Zhang, Huan Sun, et~al.
\newblock {MMMU-Pro}: A more robust multi-discipline multimodal understanding
  benchmark.
\newblock In \emph{Proceedings of the 63rd Annual Meeting of the Association
  for Computational Linguistics (Volume 1: Long Papers)}, pages 15134--15186,
  2025.

\bibitem[Zhai et~al.(2023)Zhai, Mustafa, Kolesnikov, and
  Beyer]{zhai2023sigmoid}
Xiaohua Zhai, Basil Mustafa, Alexander Kolesnikov, and Lucas Beyer.
\newblock Sigmoid loss for language image pre-training.
\newblock In \emph{2023 IEEE/CVF International Conference on Computer Vision
  (ICCV)}, pages 11941--11952. IEEE, 2023.

\bibitem[Zhao et~al.(2025)Zhao, Xiong, and Kr{\"a}henb{\"u}hl]{zhao2025image}
Yue Zhao, Yuanjun Xiong, and Philipp Kr{\"a}henb{\"u}hl.
\newblock Image and video tokenization with binary spherical quantization.
\newblock In \emph{International Conference on Learning Representations},
  volume 2025, pages 90844--90868, 2025.

\bibitem[Zheng et~al.(2026)Zheng, Ma, Tong, and Xie]{zheng2026diffusion}
Boyang Zheng, Nanye Ma, Shengbang Tong, and Saining Xie.
\newblock Diffusion transformers with representation autoencoders.
\newblock In \emph{International Conference on Learning Representations},
  volume 2026, pages 35791--35820, 2026.

\bibitem[Zhou et~al.(2025)Zhou, Yu, Babu, Tirumala, Yasunaga, Shamis, Kahn, Ma,
  Zettlemoyer, and Levy]{zhou2025transfusion}
Chunting Zhou, Lili Yu, Arun Babu, Kushal Tirumala, Michihiro Yasunaga, Leonid
  Shamis, Jacob Kahn, Xuezhe Ma, Luke Zettlemoyer, and Omer Levy.
\newblock Transfusion: Predict the next token and diffuse images with one
  multi-modal model.
\newblock In \emph{International Conference on Learning Representations}, 2025.

\bibitem[Zhu et~al.(2025{\natexlab{a}})Zhu, Wang, Chen, Liu, Ye, Gu, Tian,
  Duan, Su, Shao, et~al.]{zhu2025internvl3}
Jinguo Zhu, Weiyun Wang, Zhe Chen, Zhaoyang Liu, Shenglong Ye, Lixin Gu, Hao
  Tian, Yuchen Duan, Weijie Su, Jie Shao, et~al.
\newblock {InternVL3}: Exploring advanced training and test-time recipes for
  open-source multimodal models.
\newblock \emph{arXiv preprint arXiv:2504.10479}, 2025{\natexlab{a}}.

\bibitem[Zhu et~al.(2025{\natexlab{b}})Zhu, Li, Xin, Xia, and
  Xu]{zhu2025addressing}
Yongxin Zhu, Bocheng Li, Yifei Xin, Zhihua Xia, and Linli Xu.
\newblock Addressing representation collapse in vector quantized models with
  one linear layer.
\newblock In \emph{Proceedings of the IEEE/CVF International Conference on
  Computer Vision}, pages 22968--22977, 2025{\natexlab{b}}.

\bibitem[Zou et~al.(2025)Zou, Liao, Zhang, Liu, and Wang]{zou2025omnimamba}
Jialv Zou, Bencheng Liao, Qian Zhang, Wenyu Liu, and Xinggang Wang.
\newblock {OmniMamba}: Efficient and unified multimodal understanding and
  generation via state space models.
\newblock \emph{arXiv preprint arXiv:2503.08686}, 2025.

\end{thebibliography}
\iflatexml
\bibliographystyle{plainnat}
\else
\bibliographystyle{\FVQSourceRoot assets/plainnat}
\fi

\clearpage
\appendix
\section{Mathematical Formulation of Visual Quantization}
\label{app:quantization_framework}

This section provides the complete formulation underlying the visual quantizer and expands the FVQ construction introduced in Section~\ref{sec:fvq}. We first define feature-space vector quantization and its training objective, then derive the tight-frame analysis and synthesis operators used by FVQ.

\subsection{Quantizer Training}

\paragraph{Nearest-Neighbor Quantization.}
Following Representation Autoencoders (RAEs)~\citep{zheng2026diffusion}, we quantize semantically structured features from a pretrained Vision Foundation Model (VFM) rather than pixel-space latents. This isolates discretization from the mismatch between high-frequency pixel signals and semantic features. The tokenizer is trained solely to reconstruct frozen VFM features and receives no pixel-level supervision.

Given an image $I$, the frozen VFM produces $N$ patch features,
$\mathbf{X}=f_{\mathrm{VFM}}(I)=\{\mathbf{x}_i\}_{i=1}^{N}$, with
$\mathbf{x}_i\in\mathbb{R}^{D_x}$. An input projection
$f_{\mathrm{in}}:\mathbb{R}^{D_x}\rightarrow\mathbb{R}^{D}$ maps each feature to
$\mathbf{z}_i=f_{\mathrm{in}}(\mathbf{x}_i)$. For a codebook
$\mathcal{C}=\{\mathbf{c}_j\}_{j=1}^{C}$ with
$\mathbf{c}_j\in\mathbb{R}^{D}$, vector quantization selects the nearest codeword:
\begin{equation}
    k_i
    = \arg\min_{j\in\{1,\ldots,C\}}
    \left\|\mathbf{z}_i-\mathbf{c}_j\right\|_2^2,
    \qquad
    \mathbf{q}_i=\mathbf{c}_{k_i}.
    \label{eq:feature_vq}
\end{equation}
The integer $k_i$ is the discrete visual ID supplied to the VLM, while
$\mathbf{q}_i$ is used only within tokenizer training to reconstruct the continuous target. A feature decoder
$f_{\mathrm{dec}}:\mathbb{R}^{D}\rightarrow\mathbb{R}^{D_x}$ produces
$\hat{\mathbf{x}}_i=f_{\mathrm{dec}}(\mathbf{q}_i)$.

\paragraph{Tokenizer Training Objective.}
The tokenizer is optimized using feature reconstruction, codebook, and commitment losses:
\begin{equation}
    \mathcal{L}_{\mathrm{tokenizer}}
    = \frac{1}{N}\sum_{i=1}^{N}
    \left(
    \left\|\operatorname{sg}(\mathbf{x}_i)-\hat{\mathbf{x}}_i\right\|_2^2
    + \left\|\operatorname{sg}(\mathbf{z}_i)-\mathbf{q}_i\right\|_2^2
    + \beta\left\|\mathbf{z}_i-\operatorname{sg}(\mathbf{q}_i)\right\|_2^2
    \right),
    \label{eq:tokenizer_objective}
\end{equation}
Here, $\operatorname{sg}(\cdot)$ denotes stop-gradient and $\beta$ controls the commitment strength. The reconstruction term trains the quantized representation and feature decoder to recover the frozen VFM target. The codebook term moves the selected codeword toward the encoder output, while the commitment term encourages the encoder output to remain close to its selected codeword. The feature-space objective preserves the semantic target supplied by the VFM without requiring a pixel decoder.

\subsection{Factorized Vector Quantization: Design and Derivation}
\label{app:fvq_derivation}

\paragraph{Design Requirements.}
The analysis in Section~\ref{sec:effective_bit_budget} shows that increasing the number of codebooks expands the nominal bit budget, but does not guarantee a corresponding increase in effective semantic capacity. A scalable multi-codebook quantizer must therefore satisfy three requirements. First, each codebook should receive a fixed-dimensional view derived from the full feature, so its input does not shrink as $L$ grows. Second, these views should collectively preserve the original feature before quantization, ensuring that the factorization itself does not introduce another information bottleneck. Third, all codebooks should play symmetric roles, so newly added codebooks are not restricted to progressively weaker residuals. These requirements lead to Factorized Vector Quantization (FVQ), which combines dense fixed-dimensional projections, tight-frame synthesis, and symmetric parallel codebooks. Each visual patch is represented by $L$ indices with a per-patch bit budget of $L\log_2 C$.

\paragraph{Fixed-Dimensional Global Projections.}
To keep the information available to each codebook from structurally diminishing with $L$, FVQ gives every codebook a fixed-dimensional view of the full feature. Let $\mathbf{z}_i\in\mathbb{R}^{D}$ be the input to FVQ. For each codebook $\ell\in\{1,\ldots,L\}$, a projection
$\mathbf{R}_\ell\in\mathbb{R}^{d\times D}$ produces
\begin{equation}
    \mathbf{u}_{i,\ell}=\mathbf{R}_\ell\mathbf{z}_i.
    \label{eq:app_fvq_view}
\end{equation}
Each $\mathbf{R}_\ell$ is dense over the $D$ input channels, so every view is derived from the full feature. Unlike PQ, the view dimension $d$ remains fixed as $L$ increases. Stacking the projections and their outputs gives
\begin{equation}
    \mathbf{A}
    =
    \begin{bmatrix}
        \mathbf{R}_1 \\
        \vdots \\
        \mathbf{R}_L
    \end{bmatrix}
    \in\mathbb{R}^{Ld\times D},
    \qquad
    \mathbf{u}_i
    =
    \begin{bmatrix}
        \mathbf{u}_{i,1} \\
        \vdots \\
        \mathbf{u}_{i,L}
    \end{bmatrix}
    =\mathbf{A}\mathbf{z}_i.
    \label{eq:app_fvq_analysis}
\end{equation}

\paragraph{Tight-Frame Constraint.}
Fixed-dimensional dense views alone do not guarantee that their collection preserves the input. If the projection matrices are constructed independently without a global constraint, their row spaces may overlap substantially, repeatedly encoding some feature directions while providing little coverage of others. In the extreme case, a direction in the null space of the stacked operator $\mathbf{A}$ disappears from every projected view and cannot be recovered. Even when $\mathbf{A}$ has full column rank, unequal singular values can strongly amplify some directions and attenuate others, yielding a poorly conditioned representation whose geometry is difficult to recover. The dimensional requirement $Ld\geq D$ makes full-rank recovery possible, but does not by itself guarantee uniform or stable coverage. FVQ therefore imposes the tight-frame condition
\begin{equation}
    \mathbf{A}^{\top}\mathbf{A}=\rho\mathbf{I}_{D},
    \qquad
    \rho=\frac{Ld}{D}.
    \label{eq:app_fvq_tight_frame}
\end{equation}
This is a collective constraint on the stacked operator rather than an orthogonality requirement on each projection. Individual blocks $\mathbf{R}_\ell$ need not be square, orthogonal, or invertible. Since all eigenvalues of $\mathbf{A}^{\top}\mathbf{A}$ equal $\rho$, all singular values of $\mathbf{A}$ equal $\sqrt{\rho}$. The stacked transform consequently has full column rank, contains no lost input direction, and preserves energy uniformly:
\begin{equation}
    \left\|\mathbf{A}\mathbf{z}_i\right\|_2^2
    =\mathbf{z}_i^{\top}\mathbf{A}^{\top}\mathbf{A}\mathbf{z}_i
    =\rho\left\|\mathbf{z}_i\right\|_2^2.
    \label{eq:app_fvq_energy}
\end{equation}
Every feature direction is therefore scaled by the same factor rather than being amplified or suppressed according to direction. The transform is well-conditioned, and its transpose yields the closed-form synthesis operator derived below. The value $\rho=Ld/D$ normalizes the aggregate transform as the number and dimension of the views change.

\paragraph{QR-Based Construction.}
We use QR decomposition as a direct and numerically stable way to construct an operator that satisfies Eq.~\eqref{eq:app_fvq_tight_frame}. We first sample a fixed Gaussian matrix
$\mathbf{G}\in\mathbb{R}^{Ld\times D}$, where $Ld\geq D$, and compute its reduced QR decomposition:
\begin{equation}
    \mathbf{G}=\mathbf{Q}\mathbf{T},
    \qquad
    \mathbf{Q}^{\top}\mathbf{Q}=\mathbf{I}_{D}.
    \label{eq:app_fvq_qr}
\end{equation}
Here, $\mathbf{Q}\in\mathbb{R}^{Ld\times D}$ contains orthonormal columns and
$\mathbf{T}\in\mathbb{R}^{D\times D}$ is upper triangular. We use $\mathbf{T}$ for the triangular factor to avoid confusing it with the projection blocks $\mathbf{R}_\ell$. Only $\mathbf{Q}$ is required after orthonormalization. We define
\begin{equation}
    \mathbf{A}=\sqrt{\rho}\,\mathbf{Q},
    \qquad
    \mathbf{A}^{\top}\mathbf{A}
    =\rho\mathbf{Q}^{\top}\mathbf{Q}
    =\rho\mathbf{I}_{D},
    \label{eq:app_fvq_qr_scaling}
\end{equation}
and partition the rows of $\mathbf{A}$ into $L$ consecutive blocks to obtain
$\{\mathbf{R}_\ell\}_{\ell=1}^{L}$. QR decomposition is therefore not an additional learned module. Its role is to turn a random dense matrix into a fixed column-orthonormal basis whose scaled form exactly satisfies the tight-frame constraint. The resulting projections remain fixed throughout training.

\paragraph{Symmetric Parallel Quantization.}
To keep all codebooks structurally equivalent as $L$ grows, FVQ quantizes the projected views independently and in parallel. Each view $\mathbf{u}_{i,\ell}$ is assigned to a codeword from
$\mathcal{C}_\ell=\{\mathbf{c}_{\ell,j}\}_{j=1}^{C}$, where
$\mathbf{c}_{\ell,j}\in\mathbb{R}^{d}$. Nearest-neighbor assignment is performed independently for every view:
\begin{equation}
    k_{i,\ell}
    =\arg\min_{j\in\{1,\ldots,C\}}
    \left\|\mathbf{u}_{i,\ell}-\mathbf{c}_{\ell,j}\right\|_2^2,
    \qquad
    \mathbf{q}_{i,\ell}=\mathbf{c}_{\ell,k_{i,\ell}}.
    \label{eq:app_fvq_quantization}
\end{equation}
The patch is represented by the tuple $(k_{i,1},\ldots,k_{i,L})$. Because all codebooks receive views of the same dimension and operate in parallel, none is conditioned on the output or residual of another. Increasing $L$ thus adds another structurally equivalent codebook and increases the nominal bit budget linearly.

\paragraph{Synthesis and Exact Recovery.}
Let
$\mathbf{q}_i=[\mathbf{q}_{i,1}^{\top},\ldots,\mathbf{q}_{i,L}^{\top}]^{\top}$
denote the stacked quantized views. Since
$\mathbf{A}^{\top}\mathbf{A}=\rho\mathbf{I}_{D}$, the scaled transpose
$\rho^{-1}\mathbf{A}^{\top}$ is a left inverse of the analysis transform. FVQ therefore synthesizes the quantized feature as
\begin{equation}
    \hat{\mathbf{z}}_i
    =\frac{1}{\rho}\mathbf{A}^{\top}\mathbf{q}_i
    =\frac{1}{\rho}\sum_{\ell=1}^{L}
    \mathbf{R}_\ell^{\top}\mathbf{q}_{i,\ell}.
    \label{eq:app_fvq_synthesis}
\end{equation}
Before quantization, substituting $\mathbf{u}_i=\mathbf{A}\mathbf{z}_i$ gives
\begin{equation}
    \frac{1}{\rho}\mathbf{A}^{\top}\mathbf{u}_i
    =\frac{1}{\rho}\mathbf{A}^{\top}\mathbf{A}\mathbf{z}_i
    =\mathbf{z}_i.
    \label{eq:app_fvq_exact_recovery}
\end{equation}
The analysis--synthesis transform is therefore exactly information-preserving before nearest-neighbor lookup. If
$\mathbf{e}_i=\mathbf{q}_i-\mathbf{u}_i$ denotes the stacked quantization error, then
\begin{equation}
    \hat{\mathbf{z}}_i-\mathbf{z}_i
    =\frac{1}{\rho}\mathbf{A}^{\top}\mathbf{e}_i,
    \qquad
    \left\|\hat{\mathbf{z}}_i-\mathbf{z}_i\right\|_2^2
    \leq\frac{1}{\rho}\left\|\mathbf{e}_i\right\|_2^2.
    \label{eq:app_fvq_error}
\end{equation}
Thus, the factorization itself introduces no reconstruction error. Any discrepancy in $\hat{\mathbf{z}}_i$ arises from quantizing the projected views.

\paragraph{FVQ Training Objective.}
The synthesized feature is decoded as
$\hat{\mathbf{x}}_i=f_{\mathrm{dec}}(\hat{\mathbf{z}}_i)$. The single-codebook objective in Eq.~\eqref{eq:tokenizer_objective} extends to FVQ by applying the codebook and commitment terms to every projected view:
\begin{equation}
    \begin{aligned}
    \mathcal{L}_{\mathrm{FVQ}}
    ={}&\frac{1}{N}\sum_{i=1}^{N}
    \left\|\operatorname{sg}(\mathbf{x}_i)-\hat{\mathbf{x}}_i\right\|_2^2\\
    &+\frac{1}{NL}\sum_{i=1}^{N}\sum_{\ell=1}^{L}
    \left(
    \left\|\operatorname{sg}(\mathbf{u}_{i,\ell})-\mathbf{q}_{i,\ell}\right\|_2^2
    +\beta\left\|\mathbf{u}_{i,\ell}-\operatorname{sg}(\mathbf{q}_{i,\ell})\right\|_2^2
    \right).
    \end{aligned}
    \label{eq:app_fvq_objective}
\end{equation}
The reconstruction term supervises the joint contribution of all codebooks after synthesis, while the remaining terms train each codebook in its corresponding projected space. This preserves the structural symmetry of the codebooks throughout tokenizer training.

\paragraph{Design Summary.}
FVQ directly realizes the three requirements above. Dense projections provide every codebook with a fixed-dimensional view of the full feature, the tight-frame analysis--synthesis pair preserves these views collectively before quantization, and parallel independent lookup gives all codebooks symmetric roles. Together, these properties allow the nominal bit budget to grow with $L$ without structurally reducing the information available to newly added codebooks.

\section{Related Work}

\subsection{Native and Unified Multimodal Modeling}

Multimodal foundation models are moving beyond the conventional paradigm of attaching a vision encoder and lightweight adapter to a pretrained language model. Recent work pursues more native multimodal modeling along two complementary directions. The first adopts early-fusion pretraining, introducing multimodal data from the outset of language-model pretraining, as exemplified by Qwen3.5~\citep{qwen3.5} and Kimi K3~\citep{team2026kimi}. The second integrates visual processing directly into the multimodal backbone, reducing the architectural separation between vision and language. EVE~\citep{diao2024unveiling}, EVEv2~\citep{diao2025evev2}, Mono-InternVL~\citep{luo2025monointernvl}, and SenseNova-U1~\citep{diao2026sensenova} pursue encoder-free or monolithic designs, while TML-Interaction~\citep{thinkingmachines2026interactionmodels} combines encoder-free processing with early-fusion pretraining for real-time multimodal interaction. These approaches pursue native multimodality by changing when multimodal learning begins or where visual processing occurs.

A complementary line of work pursues unification at the representation level. Chameleon~\citep{team2024chameleon} and Emu3~\citep{wang2024emu3} represent images as discrete tokens and model vision and language under a shared autoregressive objective. X-Omni~\citep{geng2025x} uses a shared semantic image-token space for understanding and generation, with a diffusion decoder translating predicted tokens into images. LongCat-Next~\citep{team2026longcat} lexicalizes arbitrary-resolution visual inputs through hierarchical discrete codes, while UniAR~\citep{peng2026unified} uses a shared visual tokenizer to connect multimodal understanding and generation. Beyond unifying understanding and generation, visual representations can also serve as intermediate computational states. Representation Forcing autoregressively predicts visual representations before pixels and retains them in context to guide pixel-space diffusion~\citep{wang2026representation}. Context Unrolling enables explicit reasoning across multiple modal representations before producing predictions~\citep{yang2026context}. VoT generates discrete visual-plan tokens, such as objects and layouts, to bridge high-level semantics and pixel rendering~\citep{sun2026vot}. Together, these works establish discrete visual tokens as a promising interface for native and unified multimodal models~\citep{zhou2025transfusion,chen2025janus,ma2025janusflow,deng2025emerging,xie2025show,wu2025vila,zou2025omnimamba}. Complementary work improves multimodal understanding through generative imagination for human--object interaction~\citep{yuan2026if} and tool-assisted action recognition~\citep{yuan2025video}. Our work addresses the representation bottleneck underlying this interface by studying what determines the semantic fidelity of discrete visual tokens and how that capacity can be scaled.

\subsection{Semantic Visual Representations and Tokenization}

Pretrained vision-language encoders provide the semantic feature spaces that modern visual tokenizers seek to discretize. CLIP~\citep{radford2021learning} learns transferable visual representations through contrastive language supervision, while SigLIP~\citep{zhai2023sigmoid} improves the scalability of image-text alignment through a pairwise sigmoid objective. SigLIP2~\citep{tschannen2025siglip} further strengthens semantic understanding, localization, and dense visual features. 

More recent VLM-oriented vision encoders, including Qwen3 ViT~\citep{bai2025qwen3vl}, MoonViT-v2~\citep{team2026kimi}, and InternViT~\citep{zhu2025internvl3}, are trained on large-scale task-oriented multimodal data covering OCR, document understanding, and visual question answering. Compared with CLIP-style image-text alignment, this richer supervision yields more fine-grained visual representations that better preserve textual, spatial, and structural information.

These semantically structured continuous features provide a natural target space for visual tokenization. RAE~\citep{zheng2026diffusion} uses pretrained VFM features directly as continuous latents, demonstrating their suitability for reconstruction and generation. UniTok~\citep{ma2026unitok} combines reconstruction with semantic supervision to support both understanding and generation. ViQ~\citep{yu2026viq} introduces text-aligned, position-aware quantization, while dRAE~\citep{ma2026drae} uses hyperspherical codes that better match the geometry of pretrained features.

Despite these advances, discrete visual representations continue to lag substantially behind continuous visual representations in multimodal understanding. This persistent gap motivates us to study what governs the semantic information retained when continuous features are mapped to finite discrete indices. We isolate this process within a frozen VFM feature space and use downstream multimodal understanding to evaluate the resulting representations. Our analysis shifts attention from reconstruction quality alone to the semantic information carried by discrete indices.

\subsection{Quantized Visual Encoders}

Vector quantization maps continuous features to entries in a finite codebook~\citep{oord2017neural}. Recent methods improve the optimization and utilization of this discrete space from different perspectives. FSQ~\citep{mentzer2024finite}, LFQ~\citep{yu2024language}, and BSQ~\citep{zhao2025image} replace unconstrained learned codebooks with structured scalar or binary codes, while SimVQ~\citep{zhu2025addressing} and IBQ~\citep{shi2025scalable} improve the learning and utilization of explicit codebooks.

A complementary line of work expands representation capacity through multiple codebooks. Residual vector quantization recursively encodes the residual left by preceding codebooks~\citep{lee2022autoregressive}, while product quantization partitions each feature into independent channel groups, as adopted by UniTok for visual generation and understanding~\citep{jegou2011product,ma2026unitok}. However, their gains can saturate as the number of codebooks grows. Our work studies this limitation in semantic visual representations and presents FVQ as one approach to sustaining scaling through fixed-dimensional full-feature views and symmetric parallel codebooks.

\section{Additional Experiments}

\subsection{Additional Experimental Details}
\label{app:experimental_settings}

Our model follows four sequential stages: tokenizer training, alignment, pre-training (PT), and supervised fine-tuning (SFT). Tokenizer training initializes from Qwen3-VL-8B ViT and optimizes the input projection, codebooks, and feature decoder while keeping the tight-frame projections fixed, reconstructing frozen visual features for 100B tokens. Alignment freezes both the tokenizer and the LLM and trains the visual token embeddings and merger for 10B tokens. PT then unfreezes the LLM and jointly optimizes it with the visual interface for 60B tokens while keeping the tokenizer frozen. SFT continues to optimize the LLM and visual interface for 30B tokens with the tokenizer frozen. All four stages use AdamW~\citep{loshchilov2017decoupled} with $\beta_1=0.9$ and $\beta_2=0.96$. Table~\ref{tab:training_settings} summarizes the remaining hyperparameters.

\begin{table}[htbp]
    \centering
    \caption{Training hyperparameter configurations for tokenizer training, alignment, pre-training (PT), and supervised fine-tuning (SFT).}
    \label{tab:training_settings}
    \footnotesize
    \setlength{\tabcolsep}{3pt}
    \renewcommand{\arraystretch}{1.12}
    \begin{tabularx}{\linewidth}{@{}l*{4}{>{\centering\arraybackslash}X}@{}}
        \toprule
        Setting & Tokenizer & Alignment & PT & SFT \\
        \midrule
        LLM LR
            & --
            & Frozen
            & $1\times10^{-5}$
            & $1\times10^{-5}$ \\
        Visual LR
            & $1\times10^{-4}$
            & $2\times10^{-4}$
            & $1\times10^{-5}$
            & $2\times10^{-6}$ \\
        LR schedule
            & Cosine
            & Cosine
            & Cosine
            & Constant \\
        Weight decay
            & 0
            & 0.05
            & 0.05
            & 0.05 \\
        Gradient norm clip
            & 1.0
            & 1.0
            & 1.0
            & 1.0 \\
        Optimizer
            & \multicolumn{4}{c}{AdamW ($\beta_1=0.9$, $\beta_2=0.96$)} \\
        Warm-up steps
            & 10K
            & 1K
            & 1K
            & 100 \\
        Training tokens
            & 100B
            & 10B
            & 60B
            & 30B \\
        \bottomrule
    \end{tabularx}
\end{table}

\subsection{Additional Quantitative Results}

\paragraph{Quantization Position.}
We compare quantization before and after the QwenViT merger, corresponding to spatial downsampling factors of 16 and 32. The former retains the original merger, whereas the latter uses the Pre-Buffer~\citep{team2026longcat}. Table~\ref{tab:ablation_spatial_compression} shows that the 16$\times$ configuration performs slightly better overall, suggesting that Pre-Buffer is not essential. Instead, discrete visual embeddings require a learnable transformation that aligns them with the text embedding space, a role already fulfilled by the QwenViT merger when quantization is applied before it.

\begin{table}[htbp]
    \centering
    \caption{Effect of quantization position and post-embedding adaptation. $16\times$ and $32\times$ denote spatial downsampling factors relative to the input image.}
    \label{tab:ablation_spatial_compression}
    \setlength{\tabcolsep}{4pt}
    \renewcommand{\arraystretch}{1.12}
    \small
    \resizebox{\linewidth}{!}{%
    \begin{tabular}{@{}cccccccccc@{}}
        \toprule
        Configuration & MathVista & RealWorldQA & MMStar & OCRBench & TextVQA & AI2D & ChartQA & DocVQA & Avg. \\
        \midrule
        $16\times$ + Merger & 44.5          & 57.6          & 44.2          & 658          & 69.7          & 79.3          & 68.2          & 81.7          & 63.9 \\
        $32\times$ + Pre-Buffer & 46.0 & 59.2 & 43.3 & 634 & 71.5 & 75.9 & 66.8 & 81.9 & 63.5 \\
        \bottomrule
    \end{tabular}}
\end{table}

\subsection{Additional Qualitative Visualizations}

\begin{figure}[H]
\iflatexml
\noindent\textbf{Dense Chinese Academic OCR}\par
\else
\begin{tcolorbox}[
    enhanced,
    title={Dense Chinese Academic OCR},
    attach boxed title to top left={xshift=8pt,yshift=-6pt},
    boxed title style={colback=black!60,colframe=black!70,boxrule=0.6pt,arc=2pt,left=7pt,right=7pt,top=2pt,bottom=2pt},
    coltitle=white,
    fonttitle=\bfseries,
    colback=white,
    colframe=black!65,
    boxrule=0.7pt,
    arc=2pt,
    left=8pt,right=8pt,top=18pt,bottom=8pt
]
\fi
\begin{minipage}[t]{0.48\linewidth}
    \vspace{0pt}
    \centering
    \includegraphics[width=\linewidth]{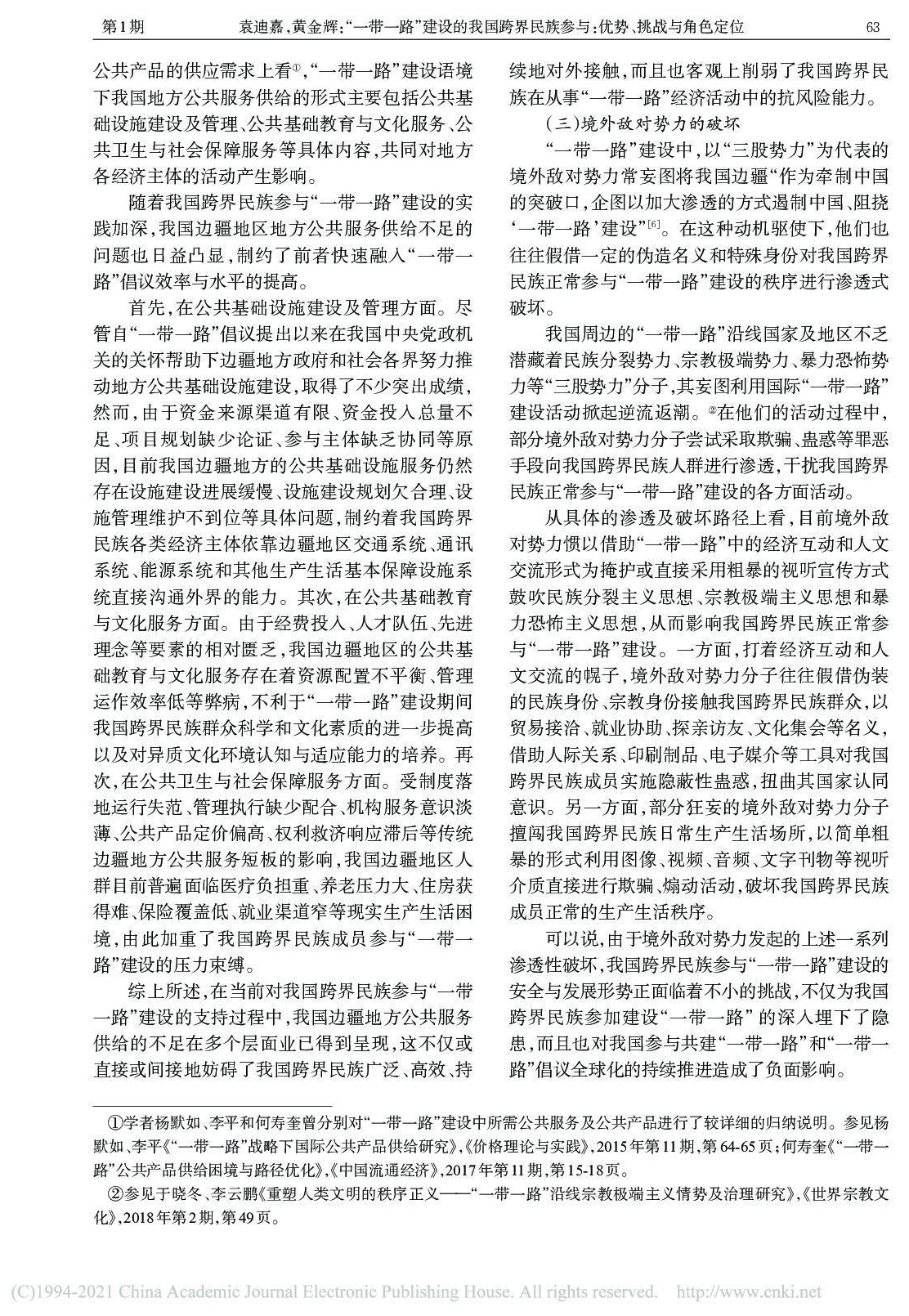}
    \vspace{3pt}

    \raggedright\fontsize{6.8}{7.8}\selectfont
    \textbf{Q:} Read all the text in the image. 

    \vspace{3pt}
    \textbf{A:} \textit{[Model output]}
    \vspace{2pt}

    \includegraphics[width=\linewidth]{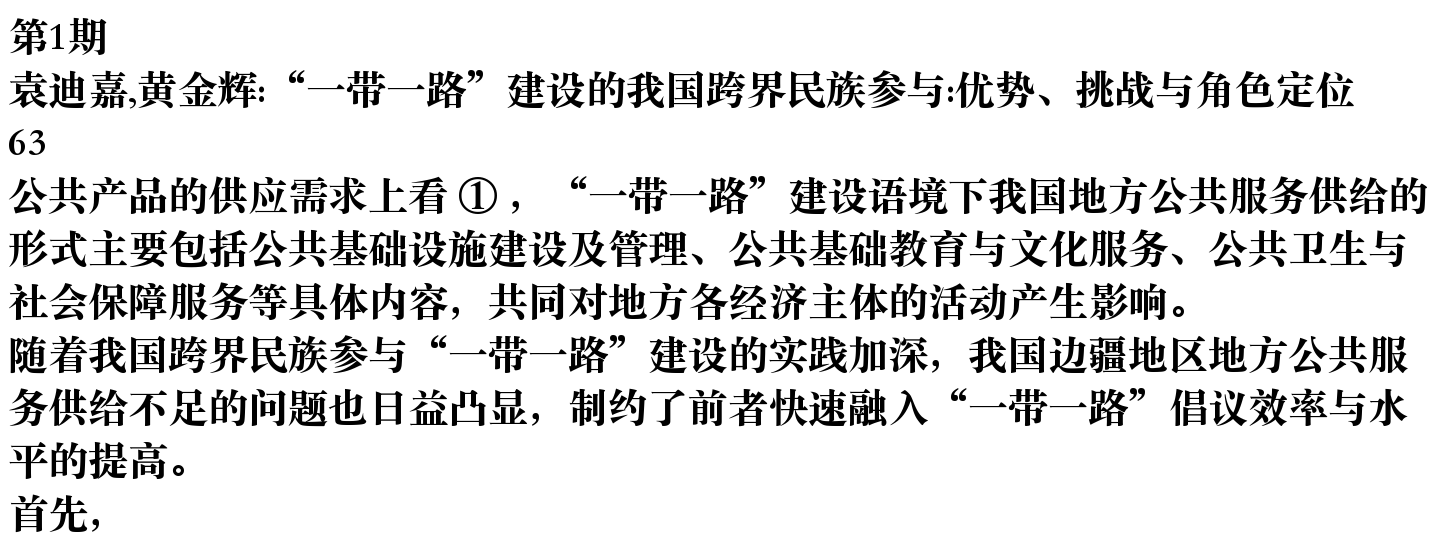}
\end{minipage}
\hfill
\begin{minipage}[t]{0.48\linewidth}
    \vspace{0pt}
    \includegraphics[width=\linewidth]{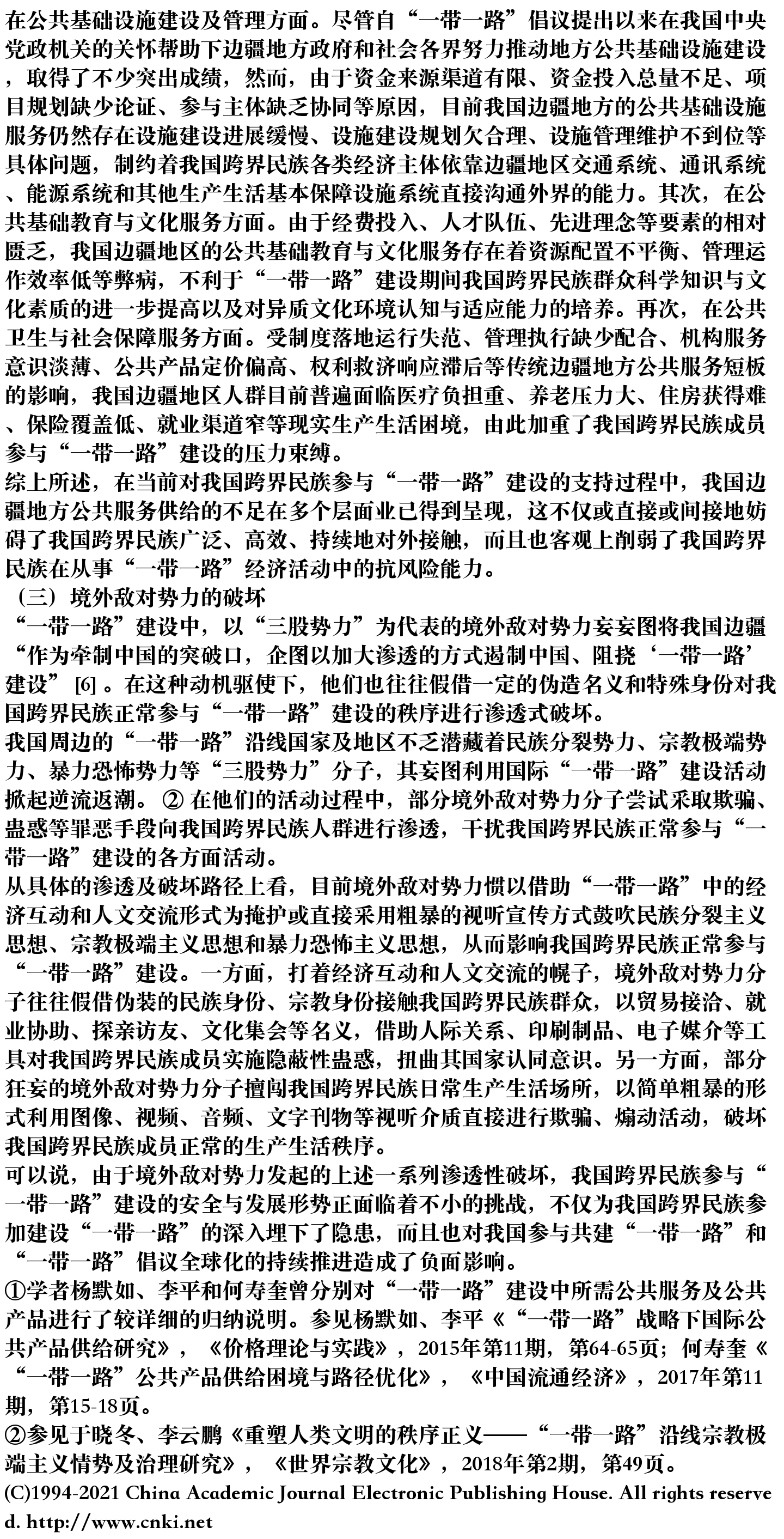}
\end{minipage}
\iflatexml\else\end{tcolorbox}\fi
\caption{Dense Chinese full-page OCR. The source page and beginning of the transcription appear on the left, with the complete output continuing on the right. The model receives an OCRBenchV2 similarity score of 0.9690.}
\label{fig:qual_dense_ocr}
\end{figure}

\begin{figure}[H]
\iflatexml
\noindent\textbf{Dense English Legal OCR}\par
\else
\begin{tcolorbox}[
    enhanced,
    title={Dense English Legal OCR},
    attach boxed title to top left={xshift=8pt,yshift=-6pt},
    boxed title style={colback=black!60,colframe=black!70,boxrule=0.6pt,arc=2pt,left=7pt,right=7pt,top=2pt,bottom=2pt},
    coltitle=white,
    fonttitle=\bfseries,
    colback=white,
    colframe=black!65,
    boxrule=0.7pt,
    arc=2pt,
    left=8pt,right=8pt,top=18pt,bottom=8pt
]
\fi
\begin{minipage}[t]{0.48\linewidth}
    \vspace{0pt}
    \centering
    \includegraphics[width=0.85\linewidth]{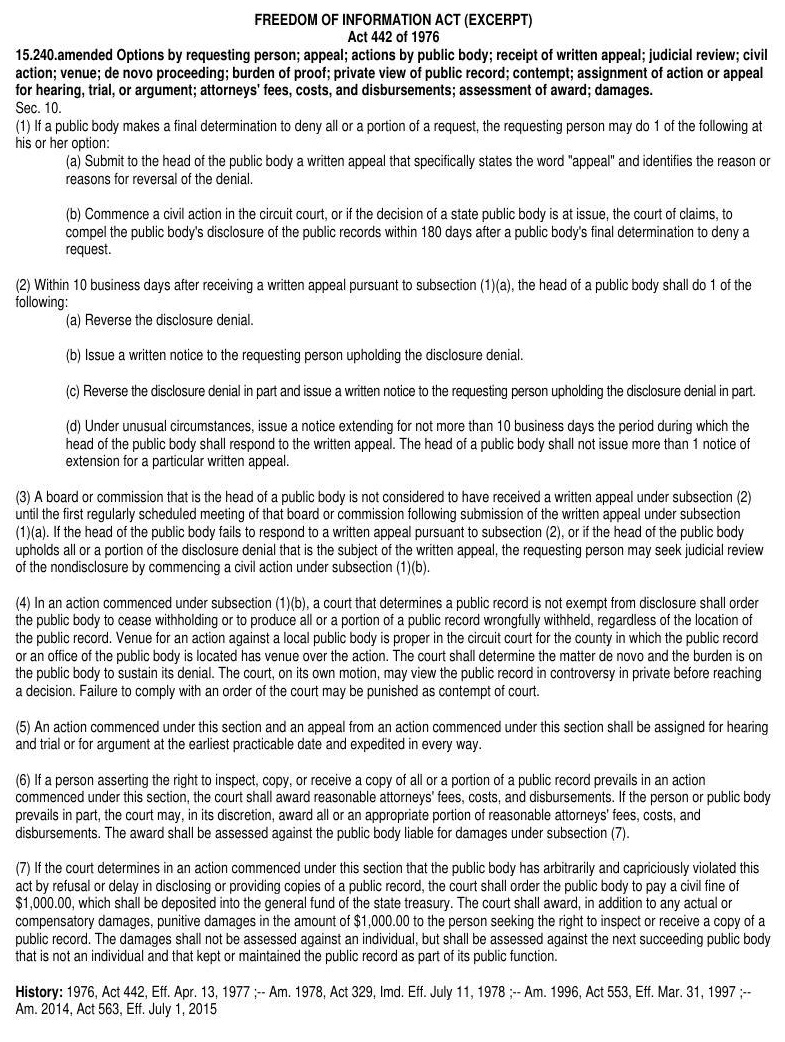}
    \vspace{3pt}

    \raggedright\fontsize{6.8}{7.8}\selectfont
    \textbf{Q:} Read all the text in the image and separate the text with spaces.

    \vspace{3pt}
    \textbf{A:} \textit{[Model output]}
    \vspace{2pt}

    \includegraphics[width=\linewidth]{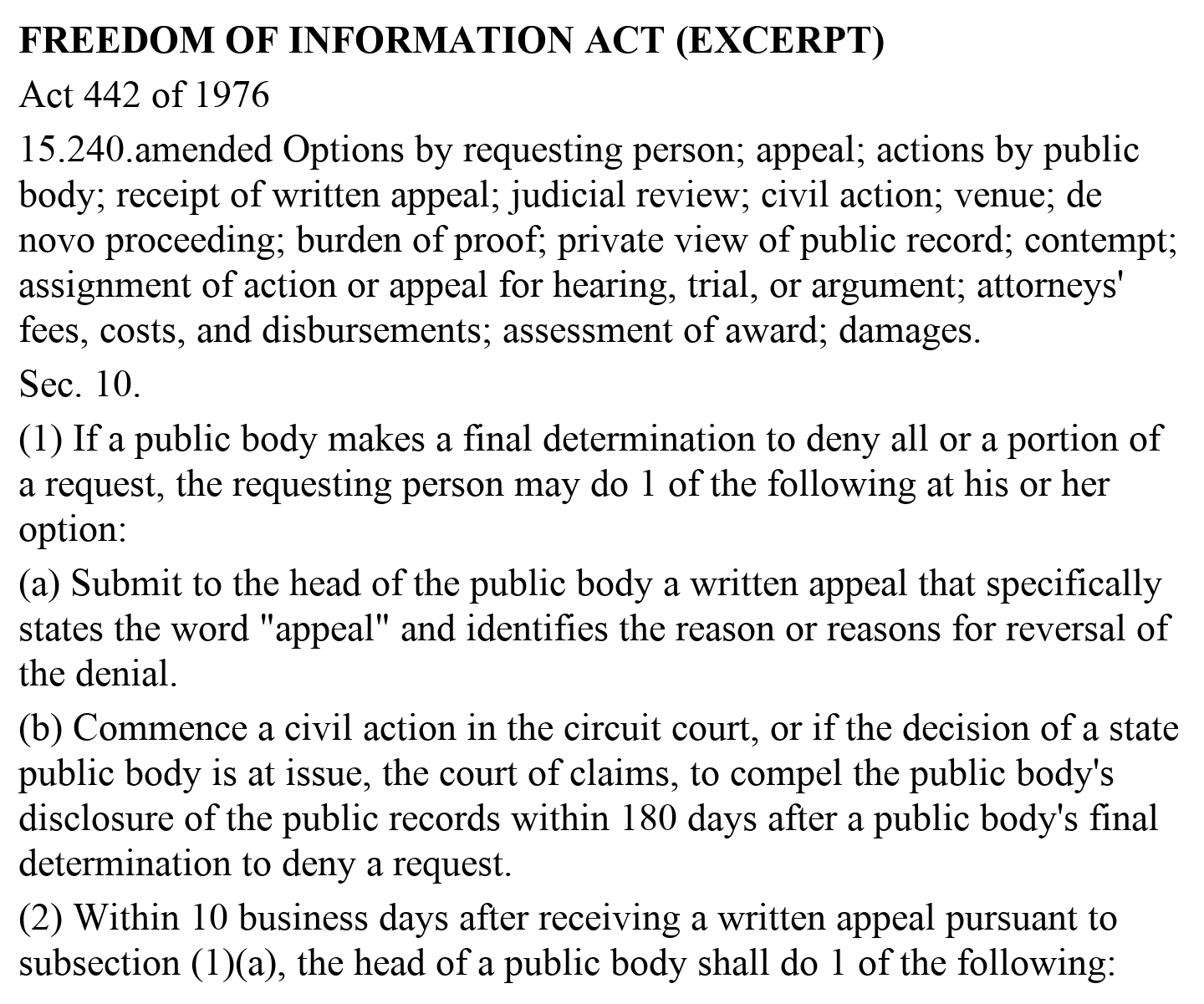}
\end{minipage}
\hfill
\begin{minipage}[t]{0.48\linewidth}
    \vspace{0pt}
    \includegraphics[width=\linewidth]{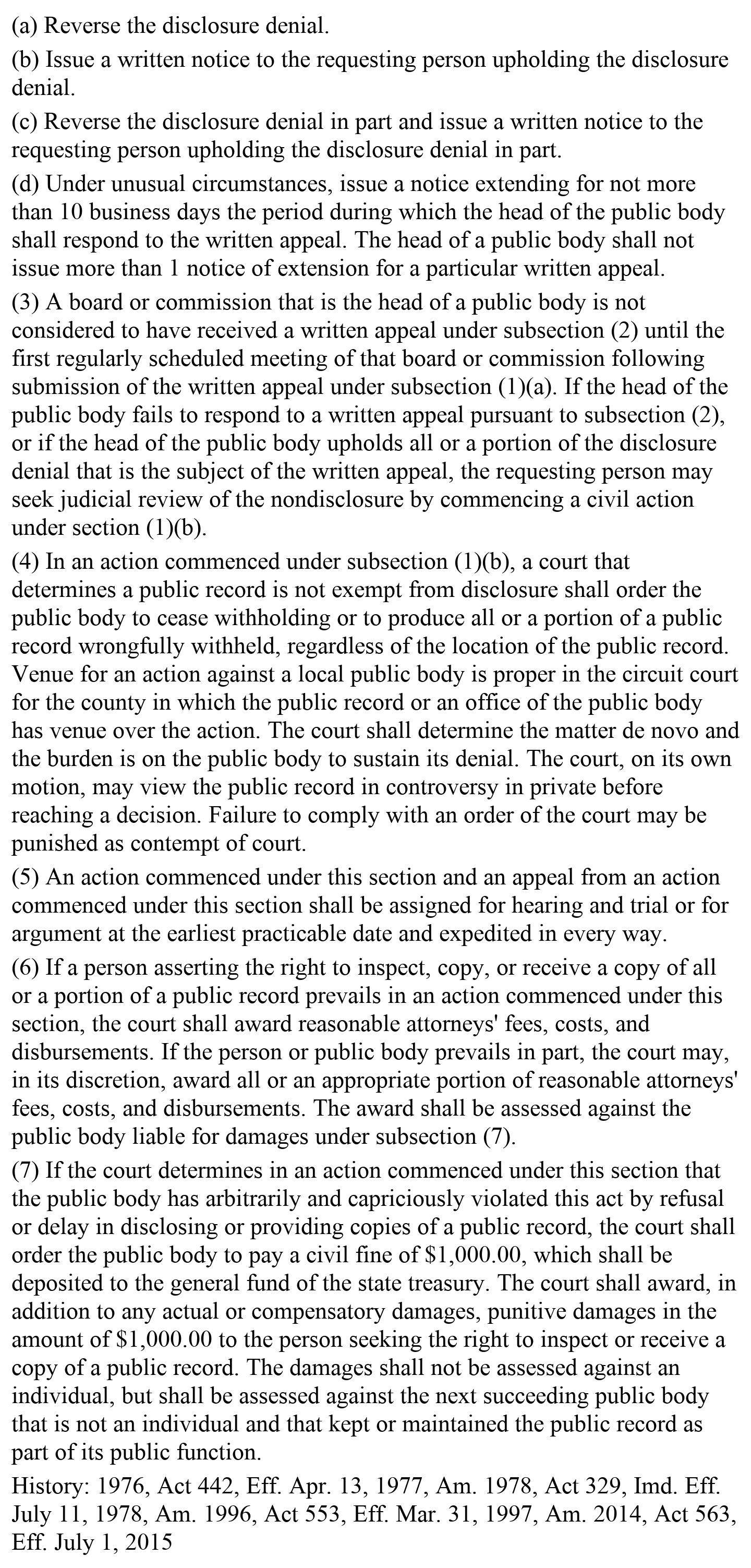}
\end{minipage}
\iflatexml\else\end{tcolorbox}\fi
\caption{Dense English full-page OCR. The source page and beginning of the transcription appear on the left, with the complete output continuing on the right. The model receives an OCRBenchV2 similarity score of 0.9828.}
\label{fig:qual_dense_ocr_en}
\end{figure}

\begin{figure}[H]
\centering
\hrule\vspace{5pt}
\textbf{Spatial Reasoning}\par\vspace{5pt}
\setlength{\abovedisplayskip}{5pt}
\setlength{\belowdisplayskip}{5pt}
\setlength{\abovedisplayshortskip}{3pt}
\setlength{\belowdisplayshortskip}{3pt}
\begin{minipage}[t]{0.30\textwidth}
\vspace{0pt}\centering
\includegraphics[width=0.96\linewidth]{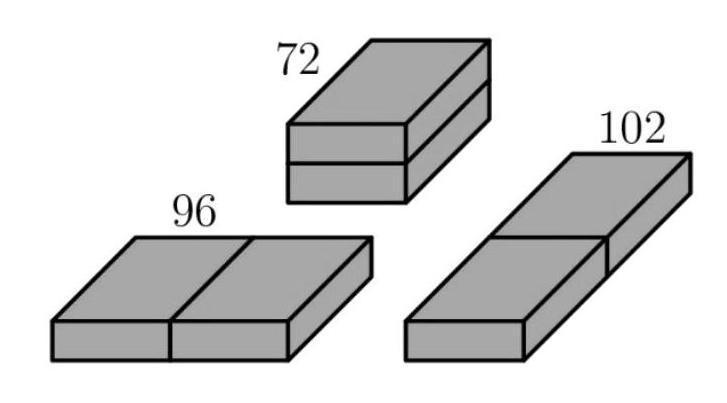}
\par\vspace{5pt}\raggedright\scriptsize
\textbf{Q:} Two identical bricks can be placed side by side in three different ways as shown in the diagrams. The surface areas of the resulting cuboids are 72, 96, and $102\,\mathrm{cm}^{2}$. What is the surface area (in $\mathrm{cm}^{2}$) of one brick?
\end{minipage}\hfill
\begin{minipage}[t]{0.64\textwidth}
\vspace{0pt}\raggedright\fontsize{7}{7.8}\selectfont
\textbf{A:} We are given three different ways to place two identical bricks side by side, forming cuboids with surface areas 72, 96, and 102 cm$^2$. We are to find the surface area of \textbf{one} brick.\par\vspace{3pt}

Let's denote the dimensions of \textbf{one brick} as length $l$, width $w$, and height $h$. Since the bricks are identical, when placed side by side, we combine them along one dimension. So the three configurations correspond to combining the bricks along length, width, or height. Let's assume the three configurations are:\par
1. Placing two bricks side by side along the \textbf{length}: so new dimensions are $2l\times w\times h$.\par
2. Placing two bricks side by side along the \textbf{width}: so new dimensions are $l\times2w\times h$.\par
3. Placing two bricks side by side along the \textbf{height}: so new dimensions are $l\times w\times2h$.\par
Each configuration gives a different surface area.\par\vspace{3pt}

The surface area of a cuboid is
\[
SA=2(lw+lh+wh).
\]
\end{minipage}

\par\vspace{6pt}
\begin{minipage}[t]{\textwidth}
\vspace{0pt}\raggedright\fontsize{7}{7.8}\selectfont
So for each configuration:\par\vspace{2pt}

\textbf{Configuration 1: $2l\times w\times h$}\par
\[
SA_1=2[(2l)(w)+(2l)(h)+(w)(h)]=2[2lw+2lh+wh]=4lw+4lh+2wh.
\]
Given: $SA_1=72$.\par\vspace{2pt}

\textbf{Configuration 2: $l\times2w\times h$}\par
\[
SA_2=2[l(2w)+l(h)+(2w)(h)]=2[2lw+lh+2wh]=4lw+2lh+4wh.
\]
Given: $SA_2=96$.\par\vspace{2pt}

\textbf{Configuration 3: $l\times w\times2h$}\par
\[
SA_3=2[lw+l(2h)+w(2h)]=2[lw+2lh+2wh]=2lw+4lh+4wh.
\]
Given: $SA_3=102$.
\end{minipage}

\par\vspace{5pt}
\begin{minipage}[t]{0.47\textwidth}
\vspace{0pt}\raggedright\fontsize{7}{7.8}\selectfont
Now we have three equations:
\[
\begin{aligned}
4lw+4lh+2wh&=72,\\
4lw+2lh+4wh&=96,\\
2lw+4lh+4wh&=102.
\end{aligned}
\]
Let's simplify these equations. Divide Equation (1) by 2:
\[
2lw+2lh+wh=36. \tag{1a}
\]
Divide Equation (2) by 2:
\[
2lw+lh+2wh=48. \tag{2a}
\]
Equation (3) remains:
\[
2lw+4lh+4wh=102. \tag{3}
\]
Now subtract (1a) from (2a):
\[
\begin{aligned}
(2lw+lh+2wh)&-(2lw+2lh+wh)=48-36,\\
-lh+wh&=12,\\
h(w-l)&=12. \quad\text{(A)}
\end{aligned}
\]
Now subtract (1a) from (3):
\[
\begin{aligned}
(2lw+4lh+4wh)&-(2lw+2lh+wh)=102-36,\\
2lh+3wh&=66. \quad\text{(B)}
\end{aligned}
\]
From Equation (A), $h(w-l)=12$. Let's solve for $h$: $h=\frac{12}{w-l}$. Plug into Equation (B):
\[
2lh+3wh=66.
\]
Factor $h$:
\[
h(2l+3w)=66.
\]
Substitute $h=\frac{12}{w-l}$:
\[
\frac{12}{w-l}(2l+3w)=66.
\]
\end{minipage}\hfill
\begin{minipage}[t]{0.47\textwidth}
\vspace{0pt}\raggedright\fontsize{7}{7.8}\selectfont
Multiply both sides by $w-l$:
\[
12(2l+3w)=66(w-l).
\]
Divide both sides by 6:
\[
2(2l+3w)=11(w-l).
\]
Expand: the left side is $4l+6w$, and the right side is $11w-11l$. Bring all terms to the left:
\[
4l+6w-11w+11l=0
\Rightarrow15l-5w=0
\Rightarrow3l=w.
\]
So, $w=3l$. Now plug back into Equation (A), $h(w-l)=12$:
\[
h(3l-l)=12
\Rightarrow h(2l)=12
\Rightarrow h=\frac{6}{l}.
\]
Now we have $w=3l$ and $h=\frac{6}{l}$. Plug into Equation (1a), $2lw+2lh+wh=36$:
\[
\begin{aligned}
2l(3l)+2l\left(\frac{6}{l}\right)+(3l)\left(\frac{6}{l}\right)
&=6l^2+12+18\\
&=6l^2+30=36.
\end{aligned}
\]
Thus, $6l^2=6$, so $l^2=1$ and $l=1$ (since length $>0$). Then $w=3l=3$ and $h=\frac{6}{l}=6$.\par\vspace{3pt}

So the dimensions of \textbf{one brick} are $l=1$, $w=3$, and $h=6$. Now compute its surface area:
\[
\begin{aligned}
SA&=2(lw+lh+wh)\\
&=2(1\cdot3+1\cdot6+3\cdot6)\\
&=2(3+6+18)=2(27)=54.
\end{aligned}
\]
\textbf{Final Answer: $\boxed{54}$}
\end{minipage}
\par\vspace{5pt}\hrule
\caption{Spatial reasoning with the complete model response, reformatted for typesetting.}
\label{fig:qual_spatial_reasoning}
\end{figure}

Figures~\ref{fig:qual_dense_ocr}--\ref{fig:qual_spatial_reasoning} present manually selected examples spanning dense OCR, long-document retrieval, fine-grained chart comparison, and visual reasoning. They illustrate the textual and reasoning-relevant information retained by the discrete representations, but are not intended as statistical evidence.

\clearpage

\clearpage

\clearpage
\begingroup
\setlength{\intextsep}{6pt}
\begin{figure}[H]
\centering
\hrule\vspace{5pt}
\begin{minipage}[t]{0.47\textwidth}
\centering\textbf{Extreme-Resolution Retrieval}\par
{\scriptsize $7035\times9660$ pixels\par}\vspace{3pt}
\includegraphics[height=0.20\textheight]{fig/qualitative/docvqa_55935_full_cropped.png}\\[-1pt]
\includegraphics[width=0.90\linewidth]{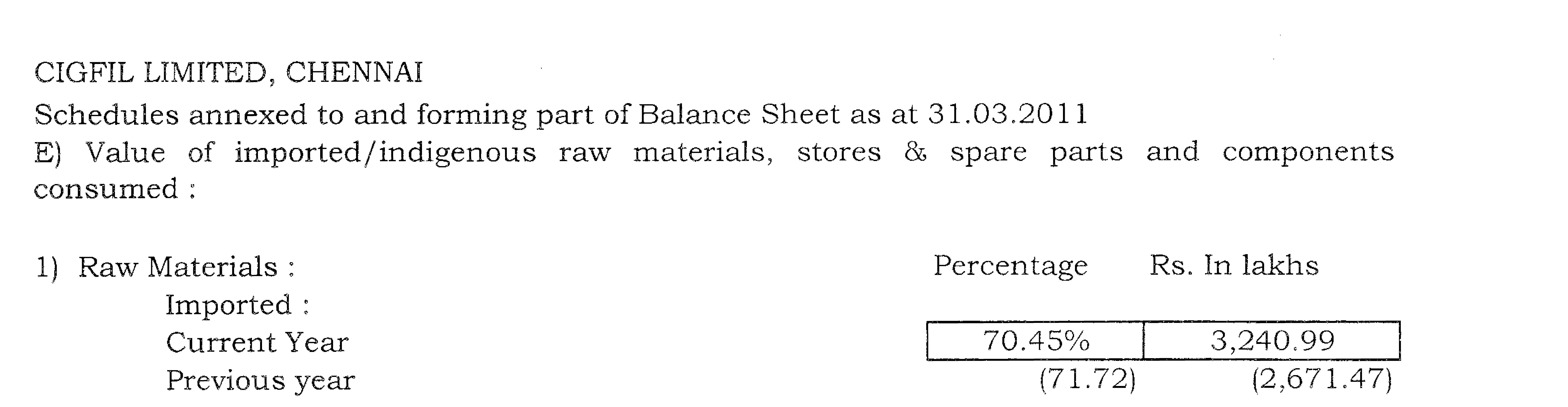}
\par\vspace{5pt}\raggedright\footnotesize
\textbf{Q:} What is the current-year percentage of raw-material import?\par\vspace{3pt}
\textbf{A:} 70.45\%.
\end{minipage}\hfill
\begin{minipage}[t]{0.47\textwidth}
\centering\textbf{Dense Plot Comparison}\par\vspace{4pt}
\includegraphics[width=0.88\linewidth]{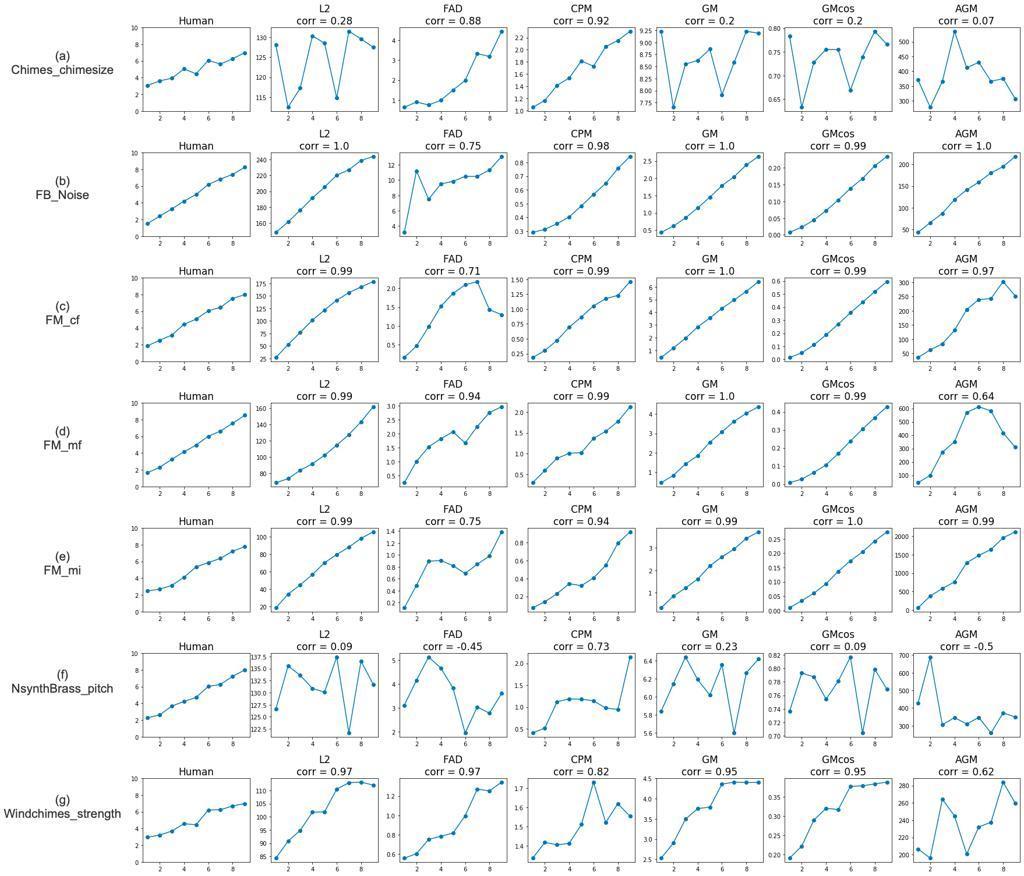}
\par\vspace{5pt}\raggedright\footnotesize
\textbf{Q:} In the \texttt{FM\_cf} row, which subplot shows the lowest correlation?\par\vspace{3pt}
\textbf{A:} In the \texttt{FM\_cf} row (labeled as (c)), the subplot that shows the lowest correlation is \textbf{FAD}, with a correlation value of \textbf{0.71}.\par\vspace{3pt}
This is explicitly written in the chart as ``FAD corr = 0.71'' in the third column of the (c) row.\par\vspace{3pt}
\textbf{Answer: FAD}
\end{minipage}
\par\vspace{5pt}\hrule
\caption{Qualitative examples of extreme-resolution document retrieval and dense plot comparison. The complete model responses are shown with formatting adapted for typesetting.}
\label{fig:qual_document_plot}
\end{figure}

\begin{figure}[H]
\centering
\hrule\vspace{5pt}
\textbf{Geometry Reasoning}\par\vspace{5pt}
\setlength{\abovedisplayskip}{5pt}
\setlength{\belowdisplayskip}{5pt}
\setlength{\abovedisplayshortskip}{3pt}
\setlength{\belowdisplayshortskip}{3pt}
\begin{minipage}[t]{0.30\textwidth}
\vspace{0pt}\centering
\includegraphics[width=0.96\linewidth]{fig/qualitative/mmmu_pro_244.png}
\par\vspace{5pt}\raggedright\scriptsize
\textbf{Q:} In circle $O$ shown below, diameter $\overline{DB}$ is perpendicular to chord $\overline{AC}$ at $E$. If $DB=34$, $AC=30$, and $DE>BE$, what is the length of $\overline{BE}$?
\par\vspace{4pt}
\textbf{Options:} A. 14; B. 8; C. 25; D. 7; E. 12; F. 10; G. 15; H. 11; I. 16; J. 9.
\end{minipage}\hfill
\begin{minipage}[t]{0.66\textwidth}
\vspace{0pt}\raggedright\fontsize{7}{7.8}\selectfont
\textbf{A:} We are given a circle with center $O$, diameter $\overline{DB}=34$, and chord $\overline{AC}=30$, which is perpendicular to diameter $\overline{DB}$ at point $E$. We are told $DE>BE$, and we are to find the length of $\overline{BE}$.\par\vspace{3pt}

\textbf{Step 1: Use the Perpendicular Chord Theorem}\par
When a diameter is perpendicular to a chord, it bisects the chord. So, since $\overline{DB}$ is perpendicular to $\overline{AC}$ at $E$, then $E$ is the midpoint of $\overline{AC}$. Thus, $AE=EC=\frac{30}{2}=15$.\par\vspace{3pt}

\textbf{Step 2: Use the Power of a Point (or intersecting chords theorem)}\par
The intersecting chords theorem states that if two chords intersect at a point inside the circle, then the products of the lengths of the segments of each chord are equal. Here, chords $\overline{AC}$ and $\overline{DB}$ intersect at $E$, so
\[
AE\cdot EC=DE\cdot EB.
\]
We know $AE=EC=15$, so
\[
15\cdot15=DE\cdot EB \quad\Rightarrow\quad 225=DE\cdot EB.
\]
Also, since $DB=34$, and $DB=DE+EB$, we have $DE+EB=34$. Let $EB=x$, then $DE=34-x$. Substitute into the equation:
\[
x(34-x)=225
\Rightarrow 34x-x^2=225
\Rightarrow x^2-34x+225=0.
\]

\textbf{Step 3: Solve the quadratic}\par
Use the quadratic formula:
\[
\begin{aligned}
x&=\frac{34\pm\sqrt{(-34)^2-4\cdot1\cdot225}}{2}\\
&=\frac{34\pm\sqrt{1156-900}}{2}
=\frac{34\pm\sqrt{256}}{2}
=\frac{34\pm16}{2}.
\end{aligned}
\]
So, $x=\frac{34+16}{2}=25$ or $x=\frac{34-16}{2}=9$. Thus, $EB=9$ or $EB=25$. But we are told $DE>BE$. If $EB=9$, then $DE=34-9=25$, so $DE>BE$ and the condition is satisfied. If $EB=25$, then $DE=9$, so $DE<BE$ and the condition is not satisfied. Therefore, $EB=9$.\par\vspace{3pt}

\textbf{Final Answer: J. 9}\par
\textbf{Answer: J}
\end{minipage}
\par\vspace{5pt}\hrule
\caption{Geometry reasoning with the complete model response, reformatted for typesetting.}
\label{fig:qual_geometry_reasoning}
\end{figure}
\endgroup

\clearpage

\end{document}